\documentclass[10pt,twocolumn,letterpaper]{article}

\usepackage[pagenumbers]{cvpr} 

\usepackage[dvipsnames,table]{xcolor}
\usepackage{graphicx}
\usepackage{booktabs}
\usepackage{amsmath}
\usepackage{multirow}
\usepackage{makecell}
\usepackage{csquotes}
\usepackage{wrapfig}
\usepackage{diagbox}

\usepackage[utf8]{inputenc}
\usepackage{color,soul}
\usepackage[T1]{fontenc}

\definecolor{trashbin_blue}{RGB}{23,211,253}
\definecolor{hydrant_green}{RGB}{37,253,53}
\definecolor{bench_orange}{RGB}{255,195,45}

\setul{0.5ex}{0.3ex}
\setulcolor{trashbin_blue}

\usepackage{CJKutf8}

\definecolor{mygray}{RGB}{230,230,230}

\definecolor{clipcolor}{gray}{1.0}
\newcommand{\clipcell}[1]{\cellcolor{clipcolor}{#1}}
\definecolor{openclipcolor}{gray}{1.0}
\newcommand{\openclipcell}[1]{\cellcolor{openclipcolor}{#1}}
\definecolor{evaclipcolor}{gray}{1.0}
\newcommand{\evaclipcell}[1]{\cellcolor{evaclipcolor}{#1}}
\definecolor{siglipcolor}{gray}{1.0}
\newcommand{\siglipcell}[1]{\cellcolor{siglipcolor}{#1}}
\definecolor{llavacolor}{gray}{1.0}
\newcommand{\llavacell}[1]{\cellcolor{llavacolor}{#1}}

\definecolor{sizescolor}{RGB}{194,215,255}
\newcommand{\sizescell}[1]{\cellcolor{sizescolor}{#1}}
\definecolor{sizemcolor}{RGB}{165,202,255}
\newcommand{\sizemcell}[1]{\cellcolor{sizemcolor}{#1}}
\definecolor{resscolor}{RGB}{194,255,191}
\newcommand{\resscell}[1]{\cellcolor{resscolor}{#1}}
\definecolor{resmcolor}{RGB}{152,243,165}
\newcommand{\resmcell}[1]{\cellcolor{resmcolor}{#1}}

\usepackage[most]{tcolorbox}

\newcommand{\virl}{V-\textit{IRL}\xspace}
\newcommand{\virlplace}{\textit{\virl Place}\xspace}

\colorlet{mapcolor}{ForestGreen}
\colorlet{langcolor}{RoyalBlue}
\colorlet{visioncolor}{WildStrawberry}
\colorlet{colabcolor}{YellowOrange}

\newtcbox{\cvtag}{enhanced,nobeforeafter,tcbox raise base,boxrule=0.4pt,top=0mm,bottom=0mm,
  right=0mm,left=4mm,arc=2pt,boxsep=2pt,before upper={\vphantom{dlg}},
colframe=visioncolor!50!black,coltext=visioncolor!25!black,colback=visioncolor!10!white,
  overlay={\begin{tcbclipinterior}\fill[visioncolor!80] (frame.south west)
    rectangle node[text=white,font=\sffamily\bfseries\tiny,rotate=90] {CV} ([xshift=4mm]frame.north west);\end{tcbclipinterior}}}

\newtcbox{\llmtag}{enhanced,nobeforeafter,tcbox raise base,boxrule=0.4pt,top=0mm,bottom=0mm,
  right=0mm,left=4mm,arc=2pt,boxsep=2pt,before upper={\vphantom{dlg}},
colframe=langcolor!50!black,coltext=langcolor!25!black,colback=langcolor!10!white,
  overlay={\begin{tcbclipinterior}\fill[langcolor!80] (frame.south west)
    rectangle node[text=white,font=\sffamily\bfseries\tiny,rotate=90] {LM} ([xshift=4mm]frame.north west);\end{tcbclipinterior}}}

\newtcbox{\colabtag}{enhanced,nobeforeafter,tcbox raise base,boxrule=0.4pt,top=0mm,bottom=0mm,
  right=0mm,left=4mm,arc=2pt,boxsep=2pt,before upper={\vphantom{dlg}},
colframe=colabcolor!50!black,coltext=colabcolor!25!black,colback=colabcolor!10!white,
  overlay={\begin{tcbclipinterior}\fill[colabcolor!80] (frame.south west)
    rectangle node[text=white,font=\sffamily\bfseries\tiny,rotate=90] {COL} ([xshift=4mm]frame.north west);\end{tcbclipinterior}}}

\newtcbox{\geotag}{enhanced,nobeforeafter,tcbox raise base,boxrule=0.4pt,top=0mm,bottom=0mm,
  right=0mm,left=4mm,arc=2pt,boxsep=2pt,before upper={\vphantom{dlg}},
colframe=mapcolor!50!black,coltext=mapcolor!25!black,colback=mapcolor!10!white,
    overlay={\begin{tcbclipinterior}\fill[mapcolor!80] (frame.south west)
        rectangle node[text=white,font=\sffamily\bfseries\tiny,rotate=90] {ENV} ([xshift=4mm]frame.north west);\end{tcbclipinterior}}}

\newcommand{\cvul}[1]{\setulcolor{visioncolor}\ul{#1}}
\newcommand{\llmul}[1]{\setulcolor{langcolor}\ul{#1}}
\newcommand{\colabul}[1]{\setulcolor{colabcolor}\ul{#1}}
\newcommand{\geoul}[1]{\setulcolor{mapcolor}\ul{#1}}

\tcbuselibrary{skins}

\AtBeginEnvironment{tcolorbox}{\footnotesize}

\newcommand{\fon}[1]{\fontfamily{#1}\selectfont}  

\newcommand{\charactercard}[6]{
    \begin{tcolorbox}[
        enhanced,
        colframe=Blue!25!gray!25!white,
        colback=Blue!25!gray!2!white,
        coltext=black,
        auto outer arc,
        left=0pt,
        right=0pt,
        top=0pt,
        bottom=0pt,
        box align=top,
        halign=left,
        valign=top,
        fit algorithm=fontsize*,
        fontupper=\tiny\fon{lmvtt}
        ]
        \begin{minipage}{.13\linewidth}
        \includegraphics[width=0.85\linewidth]{#1}  
    \end{minipage}%
    \begin{minipage}{.85\linewidth}
        {\scriptsize \textbf{Name:} #2\qquad\textbf{Age:} #3\qquad\textbf{Loc:} #4}\\[0.3em]  
        {\scriptsize \textbf{Bio:}} {#5}\\[0.3em]  
        {\scriptsize \textbf{Intention:}} {#6} 
        \vspace{0.125em}
    \end{minipage}
    \end{tcolorbox}
}

\newcommand{\agentbox}[4]{
    \begin{tcolorbox}[
    enhanced,
    beforeafter skip=0.5em,
    colframe=Blue!50!black!65,
    colback=Blue!25!gray!10!white,
    coltext=black,
    boxsep=2.5pt,
    arc=2pt,
    auto outer arc,
    left=0pt,
    right=0pt,
    top=0pt,
    bottom=0pt,
    middle=1pt,
    box align=center,
    halign=left,
    valign=top,
    fit fontsize macros,
    title={#1}  
    ]
    #2
    \begin{minipage}{\linewidth}
        \vspace{-1mm}
        \textbf{Task:} #3\\[-1.6em] 
        \begin{tcolorbox}[
            enhanced,
            colframe=gray!25!Blue,
            colback=Blue!10!white,
            coltext=black,
            boxsep=2.5pt,
            arc=2pt,
            auto outer arc,
            left=0pt,
            right=0pt,
            top=1pt,
            bottom=1pt,
            box align=top,
            halign=left,
            valign=top,
            ]
        {\textbf{Takeaway:} #4} 
        \end{tcolorbox}
    \end{minipage}
    \end{tcolorbox}
}

\usepackage[noorphans,vskip=1em,leftmargin=1em]{quoting}

\newcommand{\storytext}[1]{
    \vspace{-0.75em}
    \begin{quoting}
        \small\noindent\textit{\hspace{-1mm}{#1}}
    \end{quoting}
    \vspace{-0.75em}
}

\newcommand{\fancybreak}[0]{
    \noindent~\hfill\noindent\rule{0.75\linewidth}{0.25pt}~\hfill~
}

\definecolor{cvprblue}{rgb}{0.21,0.49,0.74}
\usepackage[pagebackref,breaklinks,colorlinks,citecolor=cvprblue]{hyperref}

\def\paperID{1404} 
\def\confName{CVPR}
\def\confYear{2024}

\title{\emph{\virl}: Grounding Virtual Intelligence in Real Life}

\newcommand\hku{1}
\newcommand\nyu{2}
\author{%
  Jihan Yang\textsuperscript{\hku}\footnotemark[1] \quad 
  Runyu Ding\textsuperscript{\hku} \quad
  Ellis Brown\textsuperscript{\nyu} \quad
  Xiaojuan Qi\textsuperscript{\hku} \quad
  Saining Xie\textsuperscript{\nyu} \vspace{.2em}\\
  \textsuperscript{\hku}The University of Hong Kong \quad  \textsuperscript{\nyu}New York University \vspace{.4em}\\
  \small{\url{https://virl-platform.github.io}}
}

\begin{document}

\twocolumn[{
    \renewcommand\twocolumn[1][]{#1}
    \maketitle
    \vspace*{-0.22in}
    \centering
    \captionsetup{type=figure}
    \hspace*{-1cm}
    \includegraphics[width=1.08\textwidth]{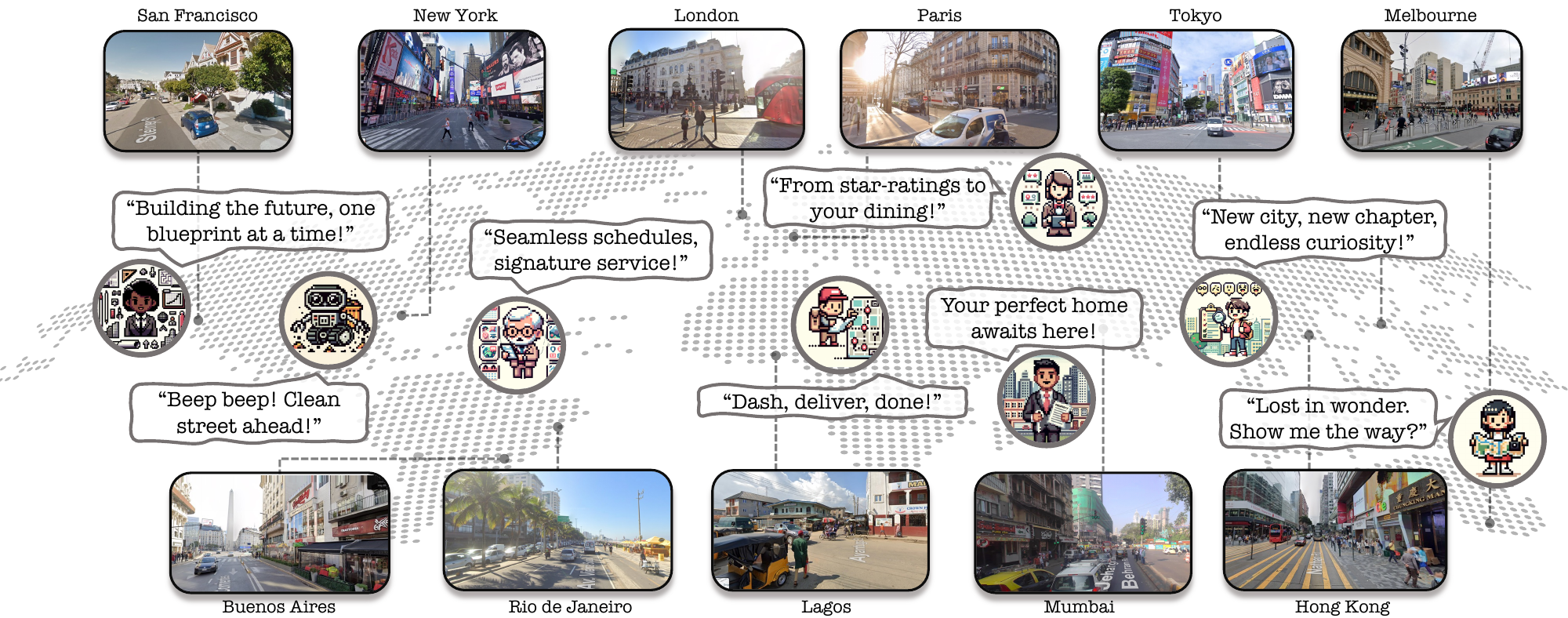}
    \vspace{-0.3cm}
    \captionof{figure}{
        \virl agents leverage real-world geospatial information and street view imagery to navigate urban terrains, execute complex tasks, and interact in real-time scenarios.
        From recommending relevant destinations to assessing city infrastructure to collaboratively giving \& following verbal directions---we develop agents that illustrate \virl's current capabilities, flexibility, and utility.
        Above all else, we present a flexible platform for researchers to harness abundant data from across the globe to create and test diverse autonomous agents.
        \\
        \\[0.15in]
    }
    \label{fig:teaser}
    \vspace{-0.5cm}
}]

\renewcommand{\thefootnote}{\fnsymbol{footnote}}
\footnotetext[1]{Work conducted during a visit to NYU.}

\begin{abstract}
\vspace{-1em}
There is a sensory gulf between the Earth that humans inhabit and the digital realms in which modern AI agents are created.
To develop AI agents that can sense, think, and act as flexibly as humans in real-world settings, it is imperative to bridge the realism gap between the digital and physical worlds.
How can we embody agents in an environment as rich and diverse as the one we inhabit, without the constraints imposed by real hardware and control?
Towards this end, we introduce \virl: a platform that enables agents to scalably interact with the real world in a virtual yet realistic environment.
Our platform serves as a playground for developing agents that can accomplish various practical tasks and as a vast testbed for measuring progress in capabilities spanning perception, decision-making, and interaction with real-world data across the entire globe.
\end{abstract}
\section{Introduction}
\label{sec:intro}
The advent of large language models (LLMs) has breathed new life into autonomous agent research by offering a universal interface for diverse capabilities, ranging from basic reasoning to complex planning and tool use~\cite{xi2023rise}.
While these developments are promising, most of these agents remain confined to text-based environments or simplistic simulations. Visual components in existing agents are either rudimentary---such as simulated tabletop environments~\cite{huang2022inner,brohan2023can}---or rely on abstracted representations using ground-truth APIs~\cite{huang2022language,wang2023voyager}.
Furthermore, the prevalent visual models employed by these agents are trained on photogenic, object-centric Internet images, which fail to capture the unpredictability and diversity of real-world scenes.

This paper aims to bridge this gap between AI agents and the sensory world by grounding them in rich, real-world environments---a crucial step towards developing autonomous agents that can effectively operate in real-life scenarios.
Our novel setting for AI agents \emph{necessitates} rich sensory grounding and perception: virtual embodiment within cities around the globe using real visual and geospatial data.

To this end, we introduce \virl, a versatile platform for building and testing virtual agents within this novel virtual-real-world setting.
\virl harnesses the power of mapping and street view data, enabling agents to navigate real-world locations, 
access up-to-date information about their surroundings, and perform practical tasks. 
With geospatial coordinates at its core, \virl is flexible and extensible, integrating with arbitrary geospatial platforms and APIs.
Moreover, \virl opens up a vast sea of visual data, allowing a simple and extensible way for researchers to evaluate vision models on realistic data distributions.

We demonstrate the versatility and adaptability of \virl by developing a series of diverse exemplar agents, each solving a unique and practical task.
As these agents hinge upon foundational language and vision models, it is critical to evaluate these models within this setting and their impact on agent performance.
We leverage the vast data available through our platform to develop \emph{global scale} benchmarks measuring the performance of underlying vision models
on images from diverse geographic and cultural contexts---evaluating their adaptability to shifting environmental, architectural, and language-specific elements.
Furthermore, we evaluate the contributions of models to agent performance on challenging tasks.
Our results illustrate the potential of \virl in bridging the gap between virtual agents and visually rich real-world environments, paving the way for future research in this direction.

In summary, our contributions are:
\begin{itemize}
    \item \textbf{\virl}: an open-source platform for building and testing agents in 
    a real-world setting that \emph{necessitates} rich sensory grounding and perception---embodiment using real geospatial data and street-view imagery.
    \item Development of \textbf{diverse exemplar agents} that showcase the platform's versatility and adaptability.
    \item \textbf{Global benchmarks} measuring the performance of foundational language and vision models (1) in isolation using our platform's real-world data and (2) on end-to-end agent performance in challenging tasks. In addition, we \textbf{discuss the robustness of ``open-world'' vision models to \emph{real-world} data from across the globe}.
\end{itemize}

\noindent We are excited to see how the research community will leverage \virl to develop and test agents that can understand and interact with the real world.
\section{Related Work}
\label{sec:related_work}
Here, we ground \virl to three streams of research.

\vspace{0.2cm}
\noindent\textbf{AI Agents.}
Agents are autonomous entities capable of perceiving their environment and acting to achieve  goals~\cite{wooldridge1995intelligent}.
Historically, agent development has leveraged symbolic and reinforcement learning methods~\cite{mnih2013playing,berner2019dota,kuttler2020nethack}, which face issues of scalability and real-world utility.
In contrast, the new wave of LLM-driven agents overcomes these challenges with text as a universal interface, enabling natural human interaction and adaptability to various tasks~\cite{wei2022chain,nakano2021webgpt,Significant_Gravitas_AutoGPT,yao2023react,shinn2023reflexion}.
Moreover, these models equip agents with complex capabilities, such as tool use and collaboration~\cite{schick2023toolformer,park2023generative,wang2023voyager,zhu2023ghost,li2023camel,wu2023autogen,hong2023metagpt}.
Yet a critical limitation persists: the agents in this new wave are entirely text-based, devoid of any tangible connection to the visual or sensory aspects of the real world.

\vspace{0.2cm}
\noindent\textbf{Embodied AI.} Embodied AI studies intelligent agents \& robots perceiving and interacting with their environment.
A significant challenge in this field is the acquisition of large quantities of realistic data. Consequently, robots are primarily trained in simulated environments~\cite{savva2019habitat,makoviychuk2021isaac,xiang2020sapien,Matterport3D,xia2018gibson} to develop skills such as navigation~\cite{anderson2018evaluation,anderson2018vision,chaplot2020object} and manipulation~\cite{gu2017deep,yenamandra2023homerobot}.
Recent advancements in LLMs~\cite{openai2023gpt,touvron2023llama,anil2023palm} have enabled embodied agents to perform long-horizon and open-end tasks in game-engines~\cite{huang2022inner,lin2023text2motion,liu2023reflect,huang2022language,shao2023lmdrive} or human rooms~\cite{liang2023code,driess2023palme,brohan2023can,huang2023voxposer,rt22023arxiv}.
However, the diversity of tasks and data is still too narrow and simplistic to enable them to operate flexibly in diverse real-world environments.

\vspace{0.2cm}
\noindent\textbf{Open-World Computer Vision.} Motivated by the success of vision-language models~\cite{radford2021learning,yuan2021florence,alayrac2022flamingo,awadalla2023openflamingo} pre-trained on large-scale web-crawled data~\cite{thomee2016yfcc100m,sharma2018conceptual,xu2023demystifying,openclip2023,schuhmann2022laionb,laurenccon2023obelisc}, open-world computer vision has received increasing attention in recent years~\cite{minderer2022simple,zhou2022detecting,li2022grounded,li2022languagedriven,ghiasi2022scaling,xu2022groupvit,li2023internet,ding2023pla,yang2024regionplc}. 
However, images and benchmarks sourced from the Internet~\cite{deng2009imagenet,kuznetsova2020open,dubey2021adaptive,asano2021pass,rojas2022dollar,li2023internet} are unavoidably biased towards specific distributions rather than truly reflecting the real \textit{world}~\cite{ramaswamy2023geode}.
Because they are trained and evaluated entirely on Internet data, existing ``open-world'' models are effectively more open-\textit{Internet} than open-\textit{world}.

\vspace{-0.1cm}
\section{Virtual Intelligence in Real Life}
\vspace{-0.1cm}
\label{sec:virl_agents}
To demonstrate the versatility of \virl, we use it to instantiate several exemplar agents in our virtual real-world environment.
In this section, we engage these agents with tasks that highlight various capabilities of our platform.
In \cref{sec:system}, we discuss the technical details of our platform and how it enables agents to interact with the real world.

For illustration, we give \virl agents character metadata, including an 8-bit avatar, a name, a short bio, and an intention they are trying to accomplish.
More concretely, agents are defined by pipelines that use this character metadata along with our platform's API and pretrained models to address complex tasks (see \cref{sec:system}).
Here we provide a high-level overview of the tasks, highlight the \virl capabilities they require, and visualize the agents solving them.

We highlight the specific \virl capabilities being employed throughout using tags and corresponding colored underlines:
\geotag{Map} $\to$
\geoul{action},
\llmtag{LLM} $\to$
\llmul{reasoning},
\cvtag{Vision} $\to$
\cvul{perception}, \&
\colabtag{Colab} $\to$
\colabul{collaboration}.

\vspace{-0.2cm}
\subsection{Earthbound Agents}
\vspace{-0.1cm}
\label{sec:virl_real_world}

\virl agents
inhabit virtual representations of real cities around the globe.
At the core of this representation are \emph{geographic coordinates} corresponding to points on the Earth's surface.
Using these coordinates, \virl allows virtual agents to \emph{ground} themselves in the real world using maps, street view imagery, information about nearby destinations, and additional data from arbitrary geospatial APIs.

\agentbox{ 
    \textbf{Route Optimizer}\hfill
    \geotag{Map}
}{ 
    \charactercard{    
        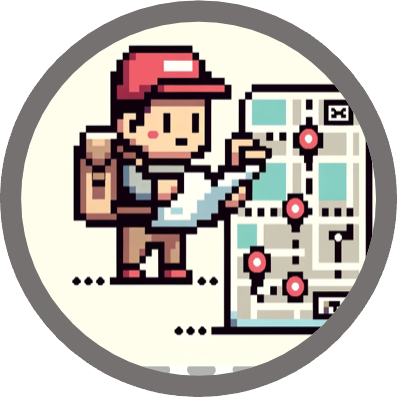
    }{Peng}{21}{NYC}{ 
        Originally from Chengdu, Sichuan, Peng is a student at PKU.
        He just arrived for a semester abroad at NYC, and is couch surfing until he gets settled.
    }{ 
        Peng needs to visit five locations around the city: his University Card Center, Residence Hall, Research Center, Library, and Student Center.
    }
}{ 
    Given a starting address and a list of waypoints, plan the shortest route to all waypoints and then follow it on street view.
}{ 
    \virl instantiates agents with \emph{real} geospatial information, and enables useful tasks like route optimization.
}
\storytext{
    Peng needs to visit several locations throughout the city to get documents signed for registration as a visiting student\ldots
}
\vspace{-0.3cm}
\fancybreak

\begin{figure}[h!]
    \centering
    \vspace{-0.4cm}
    \includegraphics[width=0.8\linewidth]{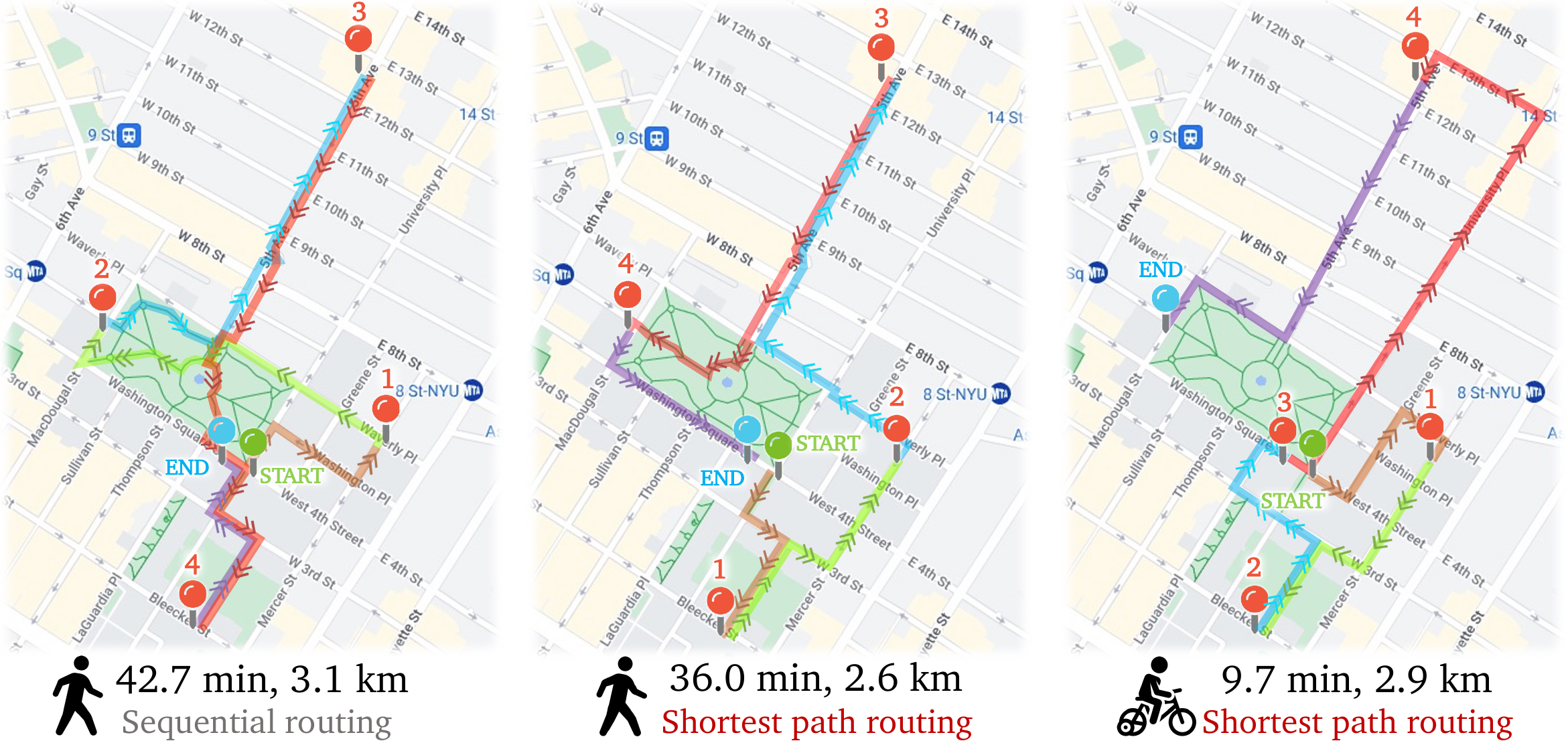}
    \vspace{-0.2cm}
    \caption{
        Finding the shortest path for Peng to travel to five places.
    }
    \vspace{-0.3cm}
    \label{fig:agent_courier}
\end{figure}

\noindent
Leveraging \geoul{Geolocation \& Mapping} capabilities,
Peng saves 7 minutes by walking along the shortest path as opposed to in order waypoint visitation as shown in \cref{fig:agent_courier}.

\vspace{-0.2cm}
\subsection{Language-Driven Agents}
\vspace{-0.1cm}
\label{sec:virl_lang_driven}

To tackle more complex tasks, we follow the pattern of language-driven agents~\cite{xi2023rise}. LLMs enable agents to flexibly reason, plan, and use external tools \& APIs.

\agentbox{ 
    \textbf{Place Recommender}\hfill
    \geotag{Map}
    \llmtag{LLM}
}{ 
    \charactercard{    
        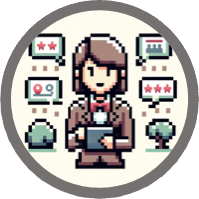
    }{Aria}{26}{NYC}{ 
        A 3rd year graduate student who loves to try new restaurants. She is always looking for new places to try, and shares her favorite spots on her blog!
    }{ 
        Pick out a lunch spot that Peng might like.
    }
    \vspace{-0.8em}
    \charactercard{    
        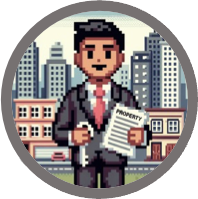
        }{Vivek}{35}{NYC}{ 
        A tech-savvy estate agent who combines his local knowledge with online tools like Zillow to find the perfect homes for his clients in the bustling city.
    }{ 
        Help Peng find a place to live for the semester.
    }
}{ 
    Given specific location, background, and intention, synthesize reviews of nearby businesses to provide a recommendation.
}{ 
    \virl exposes rich real-world information to agents that they can use for real-world tasks.
}
\storytext{
    Peng is starving for some lunch but doesn't know where to eat\dots
    Luckily, he met a nice grad student Aria during his errands who might be able to help him find a good spot\ldots
}

{
\vspace{-0.3cm}
\fancybreak
}

\begin{wrapfigure}{l}{3.55cm}
\vspace{-0.7cm}
    \begin{center}
    \includegraphics[scale=0.27]{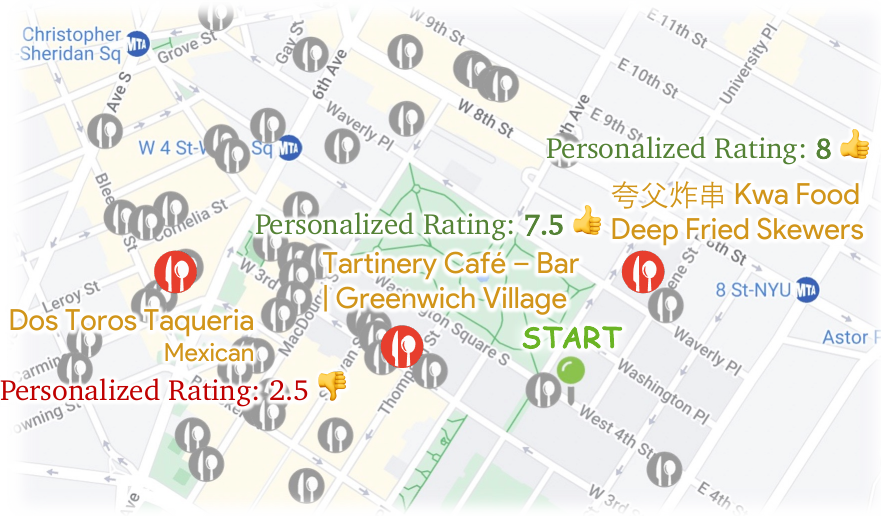}
    \end{center}
\vspace{-0.6cm}
\label{fig:agent_recommend}
\end{wrapfigure}

\noindent
Aria \geoul{searches} for possible restaurants nearby.
She then synthesizes public reviews to make final recommendations via \llmul{GPT-4}.
As Peng is new to the city and originally from Sichuan, she recommends a spicy Chinese joint \emph{Kwa Food Deep Fried Skewers} to give him a taste of home.

\storytext{
    Peng hires Vivek to help him find an apartment in East Village, Jersey City, or Long Island City for \$1k--\$3k per month close to a gym, supermarket, and public transit\ldots
}

{
\vspace{-0.2cm}
\fancybreak
}
\begin{wrapfigure}{l}{44mm}
\vspace{-0.6cm}
    \begin{center}
    \includegraphics[scale=0.34]{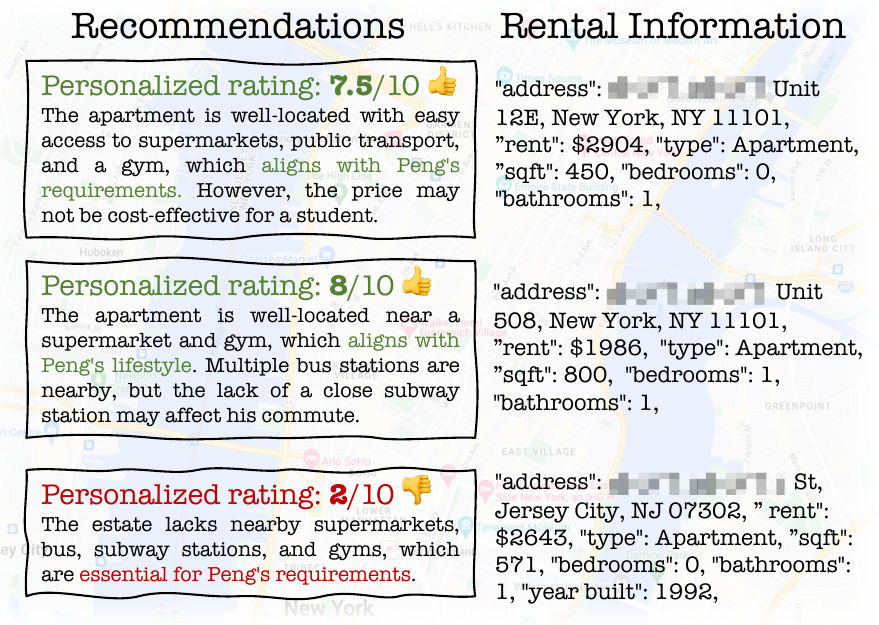}
    \end{center}
\vspace{-0.6cm}
\label{fig:agent_estate}
\end{wrapfigure}
\noindent Vivek uses real estate \llmul{APIs} to find potential apartments in Peng's desired regions and price range.
For each candidate, he researches its \geoul{proximity} to the places Peng cares about. Synthesizing these factors, Vivek provides a holistic rating and accompanying reasoning using \llmul{GPT-4}.
His top recommendation is a cost-effective 1 bedroom apartment for \$1986/mo, which is close to a supermarket, 2 bus stations, and a gym. 

\subsection{Visually Grounded Agents}
\vspace{-0.1cm}
\label{sec:virl_visual_grounding}
Although language-driven agents can address some real-world tasks using external tools, their reliance solely on text-based information limits their applicability to tasks where \emph{visual grounding} is required.
In contrast, \emph{real sensory input} is integral to many daily human activities---allowing a deep connection to and understanding of the world around us.
Agents can leverage street view imagery through the \virl platform to \emph{visually ground} themselves in the real world---opening up a wide range of \emph{perception-driven tasks}.

\begin{figure*}
    \centering
    \includegraphics[width=1\linewidth]{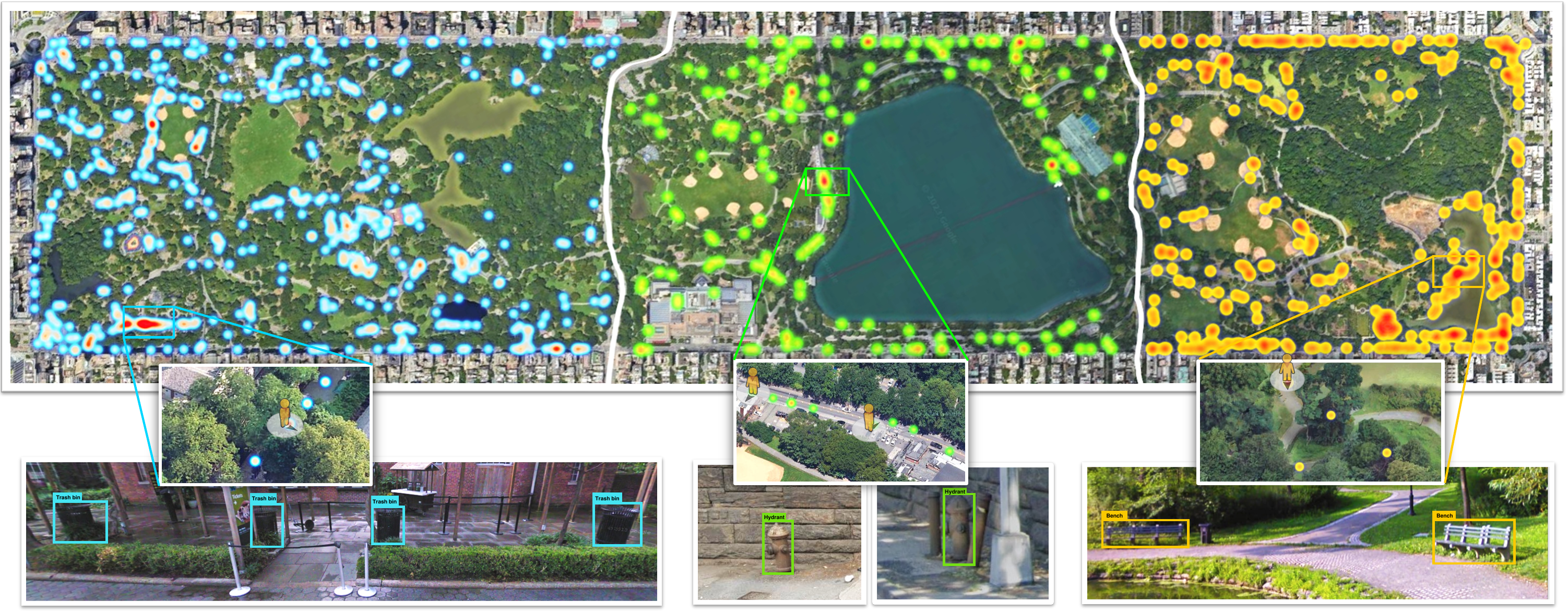}
    \vspace{-0.6cm}
    \caption{ 
    Imani's visualization of \ul{trash bins},\setulcolor{hydrant_green} \ul{fire hydrants},\setulcolor{bench_orange} \& \ul{park benches} in NYC's Central Park
    using data collected by RX-399.
    }
    \vspace{-0.2cm}
    \label{fig:agent_urban_plan}
\end{figure*}

\agentbox{ 
    \textbf{Urban Assistance Robot}\hfill
    \geotag{Map}
    \cvtag{Vision}
}{ 
    \charactercard{    
        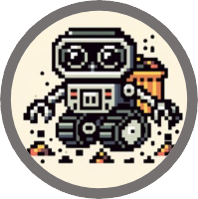
    }{RX-399}{Unk.}{HK/NYC}{ 
        This urban robot's advanced object detection, localization, and navigational telemetry systems allow it to perform perceptive tasks in busy city streets.
    }{ 
        Report the locations of trash bins to the sanitation dept.
    }
}{ 
    Travel along a specified route and detect instances of a specified object (e.g., trash bins, hydrants, benches, etc.).
}{ 
    \virl agents can use perceptive input to understand and interact with their environment.
}

\storytext{
    RX-399 is a state-of-the-art robot agent with advanced navigation and sensing capabilities.
    Its manufacturer is running a pilot program with sanitation departments in Hong Kong and New York City to assess its readiness for garbage duty\ldots
}

{
\vspace{-0.2cm}
\fancybreak
}

\begin{figure}[h!]
    \centering
    \includegraphics[width=1\linewidth]{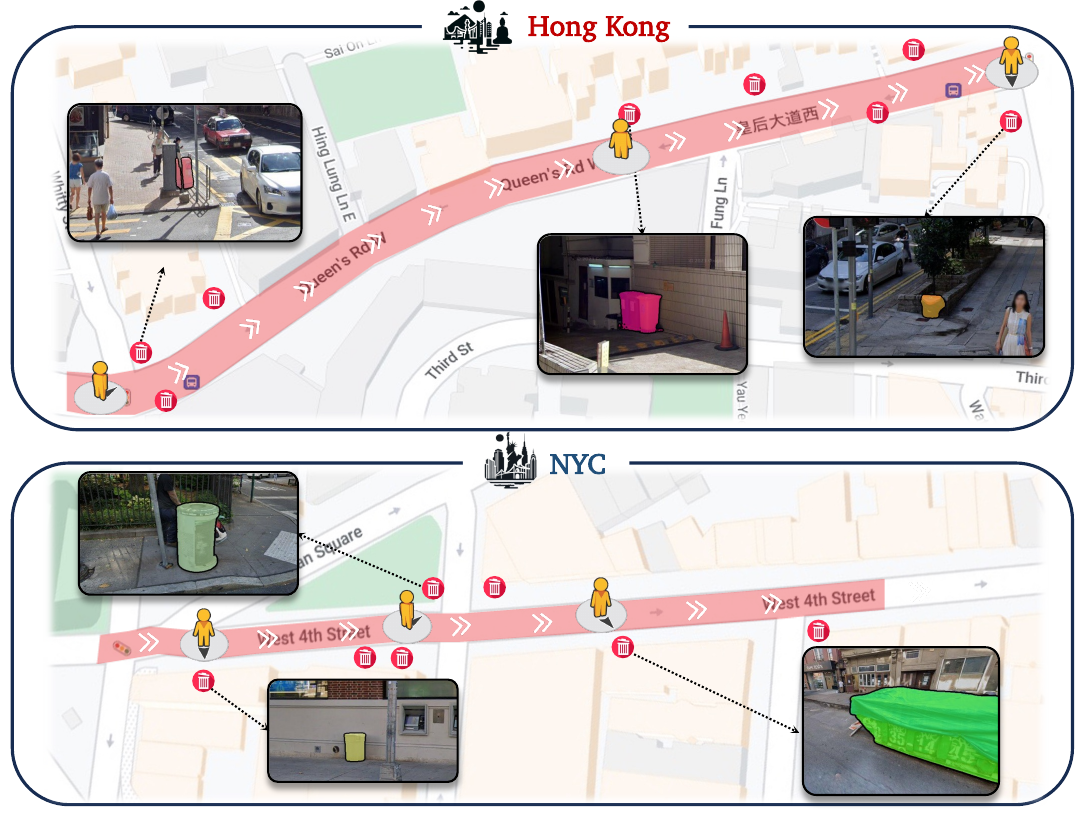}
    \vspace{-0.7cm}
    \caption{Portions of RX-399's system records in HK and NYC. 
    }
    \vspace{-0.2cm}
    \label{fig:agent_clean}
\end{figure}

\noindent
RX-399 navigates along pre-defined city routes, tagging all trash bins using its \cvul{open-world detector} and \geoul{geolocation} module as depicted in \cref{fig:agent_clean}. 
RX-399 can actively adjust its camera pose to the optimal view for each potential object thanks to our \geoul{interactive embodied environment} and the sensor-rich visual input.
\emph{During the pilot in Hong Kong, RX-399 locates eight trash bins, correctly identifying five but overlooking one. In New York, it accurately detects all five trash bins but mistakenly reports two mailboxes.}

\begin{figure}[h!]
    \centering
    \includegraphics[width=1\linewidth]{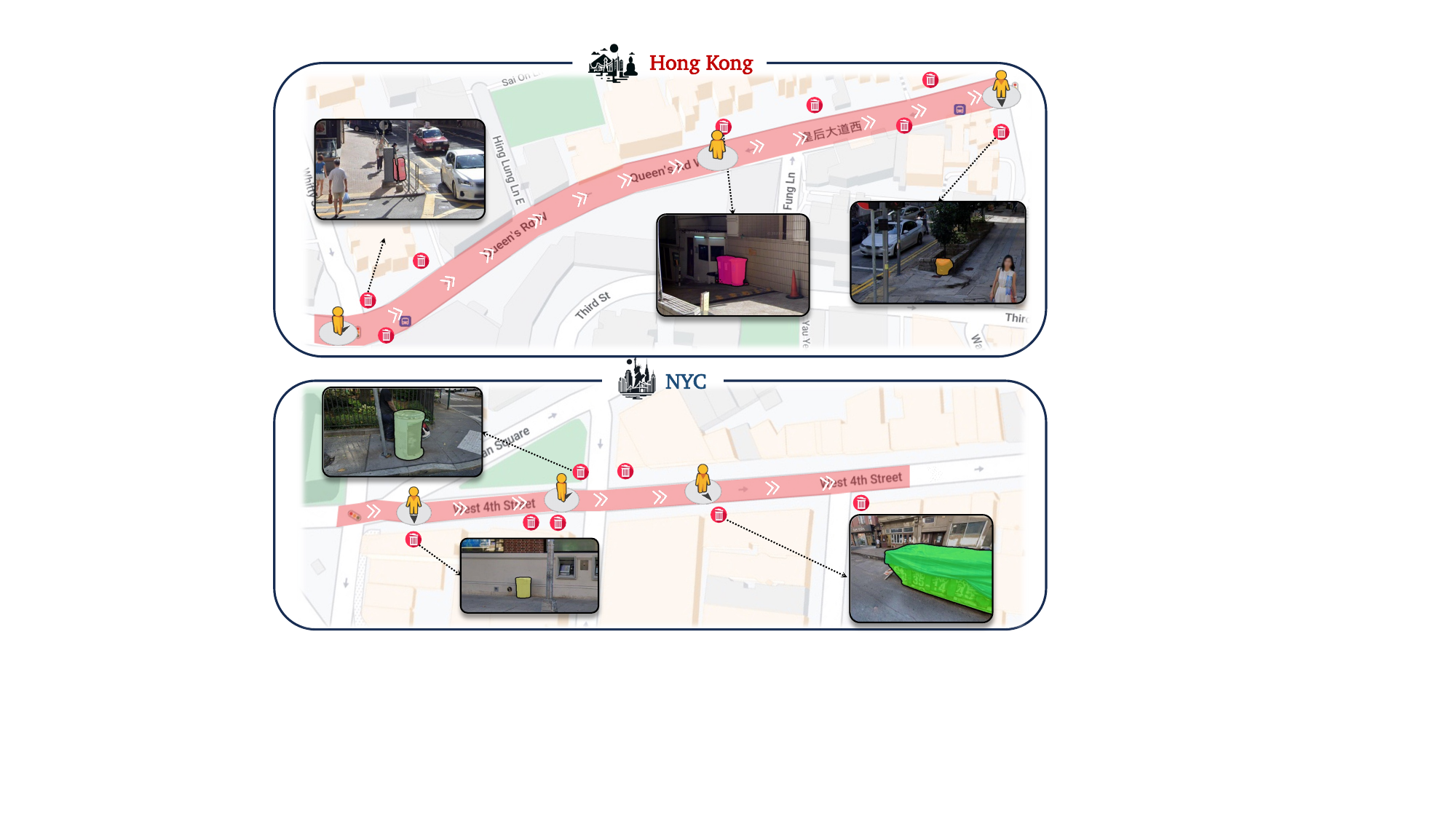}
    \vspace{-0.5cm}
    \caption{
        RX-399 avoids double-counting trash cans by identifying duplicates across different viewpoints using feature matching.
    }
    \vspace{-0.25cm}
    \label{fig:agent_clean_feature_matching}
\end{figure}

RX-399 can avoid double-counting previously seen objects by using \cvul{feature matching} to check for duplicates among prior detections (see \cref{fig:agent_clean_feature_matching}).

\agentbox{ 
    \textbf{Urban Planner}\hfill
    \geotag{Map}
    \cvtag{Vision}
}{ 
    \charactercard{    
        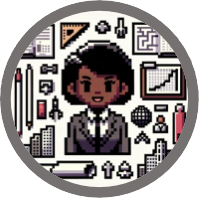
    }{Imani}{42}{NYC}{ 
        A sustainable urban development graduate, Imani is passionate about maintaining a harmonious balance between nature and urban ecosystems. 
        \vspace{0.25em}
    }{ 
        Use RX-399 to collect first-person data for her studies.
    }
}{ 
    Record the location of all instances of any specified objects (e.g., trash bins, hydrants, benches, etc.) in a specified region.
}{ 
    \virl enables realistic open-world applications requiring vast geospatial and first-person visual information.
}

\storytext{
    Imani needs to analyze the distribution of trash bins, fire hydrants, and park benches in New York's Central Park for a project with the NYC Parks \& Recreation department\ldots
}

\vspace{-0.3cm} 
\fancybreak
\vspace{-0.1cm}

\noindent
Imani sets \geoul{routes} spanning Central Park and objects of interest for RX-399, who traverses the routes and records all \cvul{detected instances}. 
After RX-399 finishes its route, Imani analyzes the collected data at different levels of detail.
As depicted in \cref{fig:agent_urban_plan}, the coarsest level shows general \geoul{distributions} of trash bins, hydrants, and benches in the park.
Imani can also zoom in to specific regions, where lighter colors represent positions with more unique instances identified.
\noindent The following table presents RX-399's counting report:

\begin{table}[h!]
\begin{center}
    \vspace{0.5em}
    \begin{small}
        \setlength\tabcolsep{3.3pt}
        \scalebox{1.0}{
        \begin{tabular}{c|ccc}
            \texttt{Category} & \texttt{Trash\hspace{0.35em}Bin} & \ \texttt{Fire\hspace{0.35em}Hydrant} \quad & \texttt{Park\hspace{0.35em}Bench$^*$}\\ 
            \hline
            \texttt{Count} & \texttt{1059} & \texttt{727} & \texttt{1015} \\
        \end{tabular}}
    \end{small}
    \caption{RX-399's counting report in Central Park, New York City. \textit{($^*$Note: contiguous benches counted as one instance).}}
\end{center}
\end{table}

\noindent
By retrieving \geoul{geotagged} sensory-rich data within RX-399, Imani can also inspect the detection results for each object to help her verify the reliability of RX-399's reports as illustrated by the bottom level in \cref{fig:agent_urban_plan}.

\agentbox{ 
    \textbf{Intentional Explorer}\hfill
    \geotag{Map}
    \llmtag{LLM}
    \cvtag{Vision}
}{ 
    \charactercard{    
        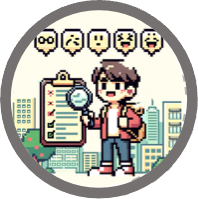
    }{Hiro}{22}{HK}{ 
        A seasoned traveler, Hiro thrives in unknown territories. He enjoys getting lost in new places instead of following the travel guide. 
    }{ 
        Hiro is looking for an authentic lunch spot that is not too spicy.}
}{ 
    Explore on foot (in street view) looking for a destination that fulfills a certain intention (e.g., lunch, shopping, etc.)
}{ 
    Agents can utilize visual detectors, VLMs and LLMs to iteratively perceive, decide, and interact in the environment.
}

\storytext{
    Hiro is starting a new journey in Hong Kong.
    He decides to explore without a specific destination in mind, looking for a good local lunch spot with food that's not too spicy\ldots
}

{
\vspace{-0.2cm}
\fancybreak
}

\begin{figure}[h!]
    \centering
    \includegraphics[width=1.05\linewidth]{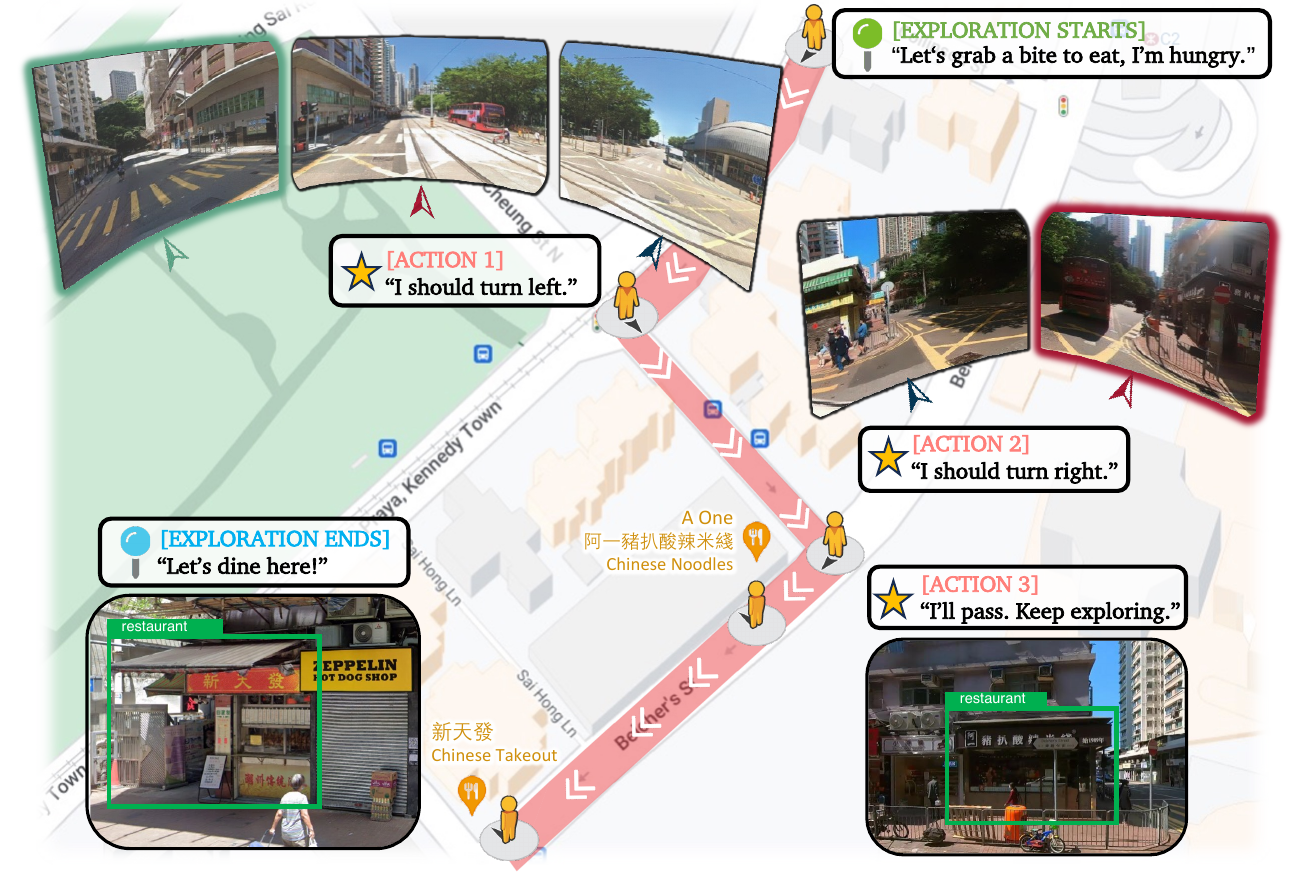}
    \caption{Visualization for Hiro's lunch exploration in HK.}
    \label{fig:agent_intention}
\end{figure}

\noindent
As depicted in \cref{fig:agent_intention}, starting at \includegraphics[height=1.0em,trim={0mm 1.5mm 0mm 1mm}]{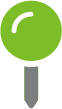}, Hiro walks down the street and encounters the first intersection. Thanks to the interactive and sensory-rich \geoul{environment}, he can adjust his pose to fetch street views for each possible path.
Using \cvul{VQA} on these views, he \llmul{decides} to turn left:

\storytext{
    \includegraphics[height=0.9em,trim={0mm 1mm 0mm 1mm}]{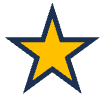} Residential buildings on the left road indicate cozy and family-run local food\ldots A better choice than the others!
}
\noindent
Then, after exploring for a block, he encounters the second intersection where he \cvul{looks around} and \llmul{decides} to turn right:
\vspace{-0.4cm}
\storytext{
    \includegraphics[height=0.9em,trim={0mm 1mm 0mm 1mm}]{figs/icons/star.png} Looks like there are some local food spots this way\ldots
}
After a few steps, Hiro finds ``\textit{A One Chinese Noodles}
\begin{CJK*}{UTF8}{bsmi}阿一豬扒酸辣米線\end{CJK*}''
using his \cvul{open-world detector}.
He \geoul{retrieves} information, ratings, and reviews for the restaurant using our platform, which \emph{connects street views to places}.
Hiro ultimately \llmul{decides} to pass on it and keep exploring because:
\vspace{-0.4cm}
\storytext{
    \includegraphics[height=0.9em,trim={0mm 1mm 0mm 1mm}]{figs/icons/star.png} Most reviews mention the \textbf{spicy} pork chop noodles\ldots 
}
Finally, at the end of the block \includegraphics[height=1.0em,trim={0mm 1.5mm 0mm 1mm}]{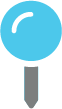}, Hiro discovers another lunch spot called ``\textit{Xintianfa} \begin{CJK*}{UTF8}{bsmi}新天發\end{CJK*}''.
He decides to dine there after \llmul{reading} numerous \geoul{online reviews} praising its authentic cuisine and diverse menu.

\subsection{Collaborative Agents}
\label{sec:virl_collaborative}

Humans often work together to solve complex real-world tasks. This collaboration promotes efficiency and effectiveness by decomposing 
a complex task into simpler sub-tasks, allowing each to be handled by an expert in its domain. 
Grounded in the world via our platform, \virl agents can leverage geospatial data and street view imagery to collaborate with other agents as well as with human users.

\subsubsection{Agent-Agent Collaboration}
\label{sec:virl_agent_agent}
As with previous agents, \colabul{collaborative} agents are designed for specific tasks; however, they can handle objectives beyond their expertise through collaboration with each other.

\vspace{0.2cm}
\agentbox{ 
    \textbf{Tourist}\hfill
    \geotag{Map}
    \llmtag{LLM}
    \cvtag{Vision}
    \colabtag{Colab}
}{ 
    \charactercard{    
        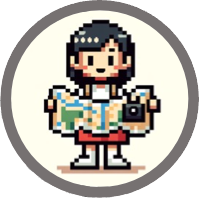
    }{Ling}{25}{NYC/SF/HK}{ 
        Ling is a spirited traveler from Taipei who is always eager to explore new cities and cultures. She is unafraid of asking locals for help when she's lost!
    }{ 
        NYC: find gifts for friends back home; go to a famous restaurant. SF: find a store to repair a broken iPhone. HK: try some authentic local food.
    }
}{
    ($i$) Ask a nearby Local agent for directions to a specific location. The Locals will preview the route on the map and in streetview and then provide walking directions in natural language, mentioning major intersections and landmarks.

    \hspace{0.8cm}
    ($ii$) Follow these directions in streetview, and if lost, ask another Local agent for assistance.
}{
    Agents can collaborate to solve complex tasks that are beyond their 
    individual expertise.
}
\vspace{0.2cm}

\storytext{
    Ling travels to cities around the world. She seeks out authentic experiences and
    is always unafraid to ask for help from Locals whenever she finds herself lost\dots
}

{
\fancybreak
}

\begin{figure*}[h!]
    \centering
    \vspace{1cm}
    \includegraphics[width=1\linewidth]{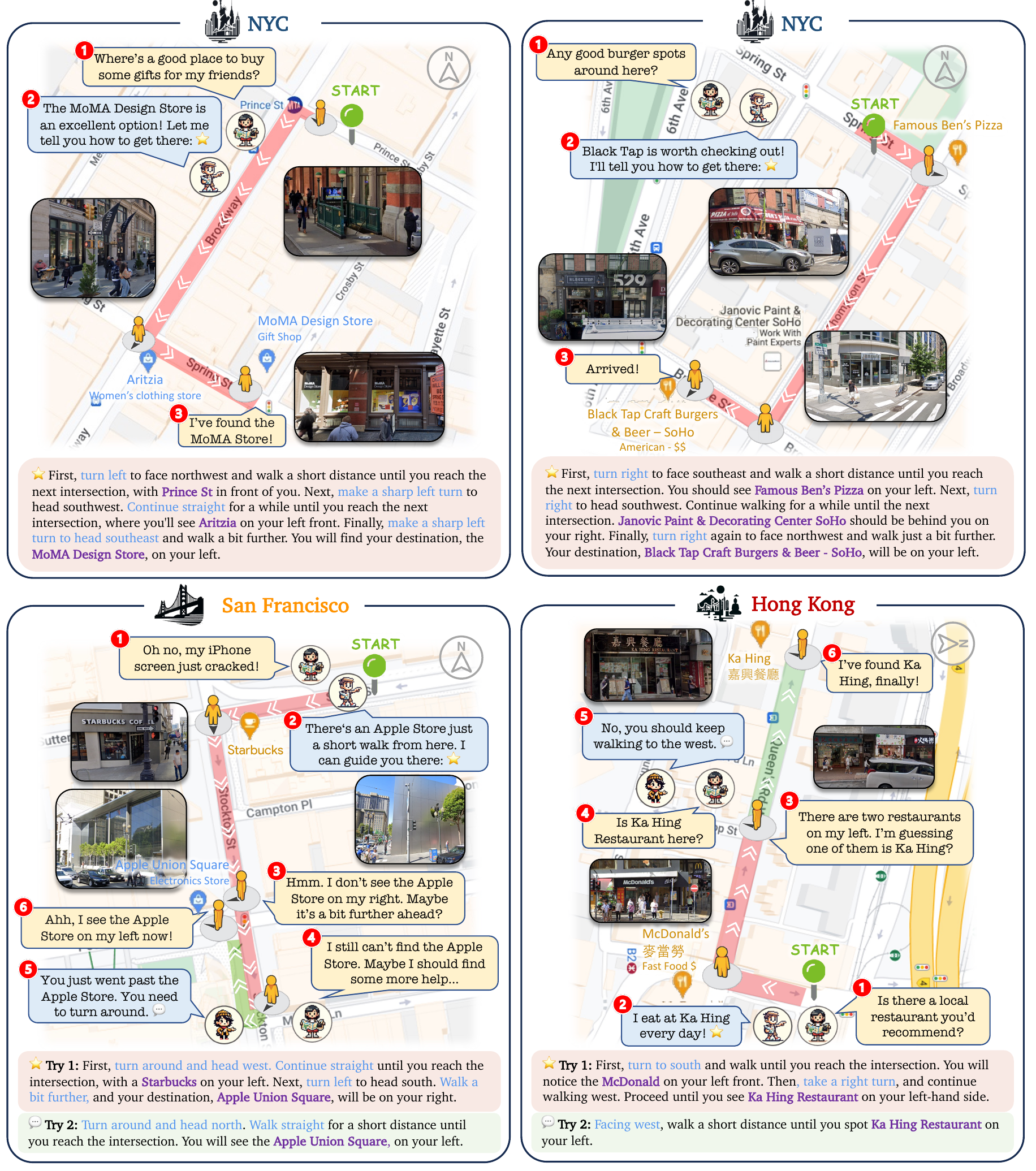}
    \vspace{-0.5cm}
    \caption{Ling and Local collaboration examples. 
    Trajectories in red and green mean Ling's first and second attempts, respectively.
    }
    \label{fig:agent_navigation}
\end{figure*}

\noindent
After obtaining route descriptions from Locals, Ling starts her journey---as shown in \cref{fig:agent_navigation}.
Grounded in our embodied platform, Ling can adjust her pose and identify visual landmarks along the streets using \cvul{open-world recognition} and her \geoul{map}. 
Correctly recognizing these landmarks helps \llmul{GPT-4} to make correct decisions about where to change direction, move forward, and stop, as seen in the top two New York City cases in \cref{fig:agent_navigation}. 
The success of these decisions made by \llmul{GPT-4} relies on the real-sensory input for visual grounding and the interactive environment from \virl.

Nevertheless, Ling may occasionally fail to find the destination.
In the bottom left San Francisco example in \cref{fig:agent_navigation}, Ling passes by the Apple Store because only its stainless steel wall is visible from her viewpoint. 
In the bottom right Hong Kong example, Ling mistakes another restaurant for her destination and stops prematurely.
Fortunately, when she makes these mistakes, Ling can ask another Local agent for new directions and start another round of navigation, which eventually leads her to the destination.

\begin{figure*}[h!]
    \centering
    \includegraphics[width=1\linewidth]{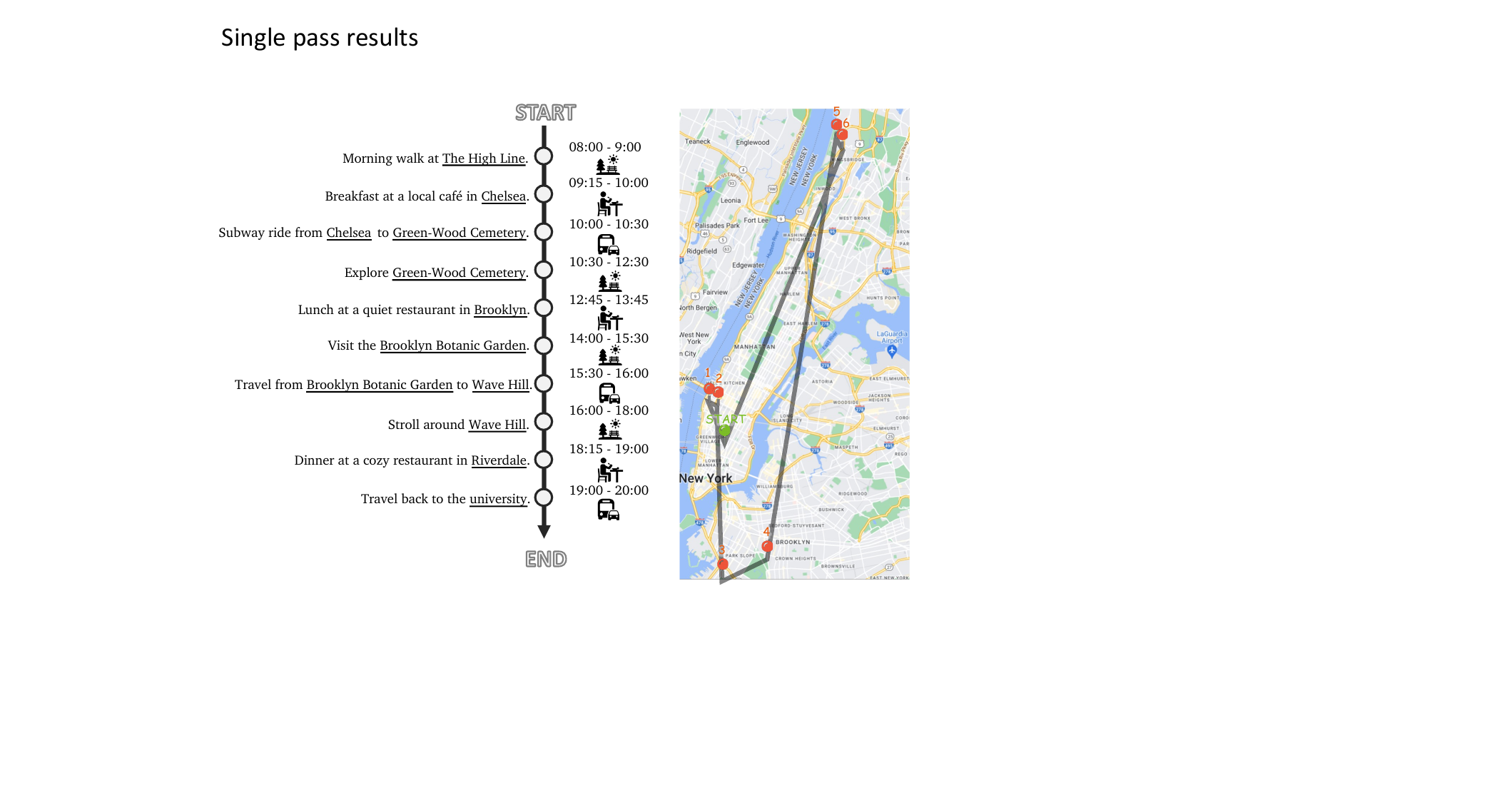}
    \vspace{-0.7cm}
    \caption{
    \emph{The Perfect Day Itinerary}: Crafted by Diego, our iterative concierge agent, this schedule is meticulously tailored, accounting for your mental and physical well-being and budget variations as your day unfolds.
    }
    \vspace{-0.2cm}
    \label{fig:agent_concierge_iterative}
\end{figure*}

\begin{figure*}[h!]
    \centering
    \includegraphics[width=1\linewidth]{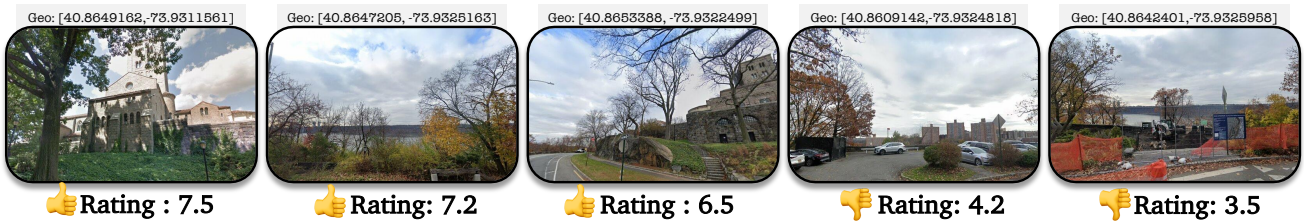}
    \vspace{-0.7cm}
    \caption{Diego traverses regions of interest to find scenic locations to add to your itinerary. 
    }
    \vspace{-0.1cm}
    \label{fig:agent_photographer}
\end{figure*}

\subsubsection{Human-Agent Collaboration}
\label{sec:virl_human_agent}
Grounded in the same environment we humans inhabit, \virl agents can collaborate with and assist real human users.

\agentbox{
    \textbf{Interactive Concierge}\hfill
    \geotag{Map}
    \llmtag{LLM}
    \cvtag{Vision}
    \colabtag{Colab}
}{
    \charactercard{    
        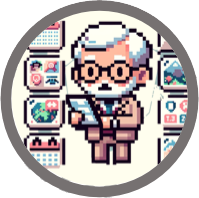
    }{Diego}{62}{NYC}{
        Diego is an expert concierge at a hotel. He's a master at creating intricate itineraries and providing valuable local advice.
    }{
        Plan personalized and practical itinerary for customer!
    }
}{
    Given a user's location, background, and intention for a day, plan a 
    full itinerary
    balancing their mental/physical state \& budget.
}{
    \virl agents can collaborate with users to solve complex tasks that require understanding the user's internal state.
}

\storytext{
As a university student in NYC,
you are excited to spend a day exploring lesser-known and tranquil places. Your friend recommended Diego, who is known for his professionalism in planning practical and personalized itineraries.
}

\vspace{-0.1cm}
\fancybreak

As depicted in \cref{fig:agent_concierge_iterative}, Diego's itinerary is tailored to 
\emph{your} (the user's)
needs. Diego not only considers your physical and mental interoception status, budget for each activity, but also anticipates your status changes and cost when you follow each event. 
He is able to take into account \emph{real} \geoul{travel times} from the \virl platform and select suitable 
destinations
by \colabul{collaborating} with another 
recommendation agent.

\begin{figure}[h!]
    \centering
    \includegraphics[width=1\linewidth]{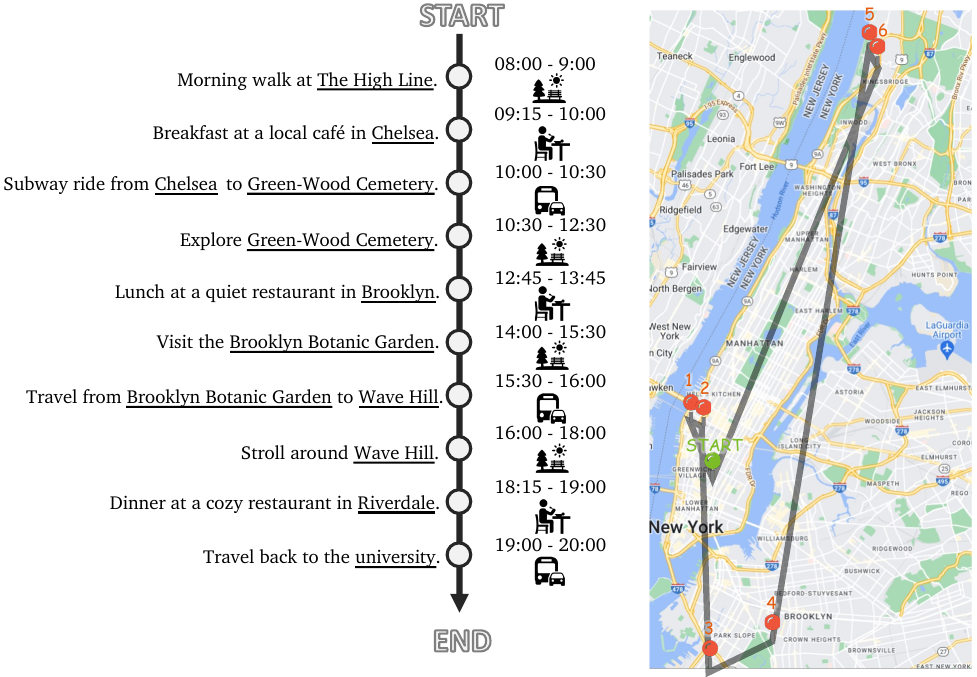}
    \vspace{-0.3cm}
    \caption{An ungrounded LLM-only concierge agent's itinerary.
    }
    \vspace{-0.1cm}
    \label{fig:agent_concierge_singlepass}
\end{figure}

In contrast, \cref{fig:agent_concierge_singlepass} shows that a simpler ``ungrounded'' LLM-only concierge agent is unable to consider the real distance and travel time between locations without access to \virl, resulting in an impractical itinerary. 
For example, lacking \geoul{real geospatial information}, the ungrounded concierge allocates only \emph{30 minutes} for travel between the ``Brooklyn Botanic Garden'' and ``Wave Hill'' in the Bronx, which actually requires \emph{60--100 minutes}\footnote{(per Google Maps \href{https://maps.app.goo.gl/SW1r5GSx3ZVo7BTr7}{https://maps.app.goo.gl/SW1r5GSx3ZVo7BTr7}).}.
The hallucinated travel times overlook geospatial realities and result in 
a plan with excessively distant destinations.

\begin{figure}[h!]
    \centering
    \includegraphics[width=1\linewidth]{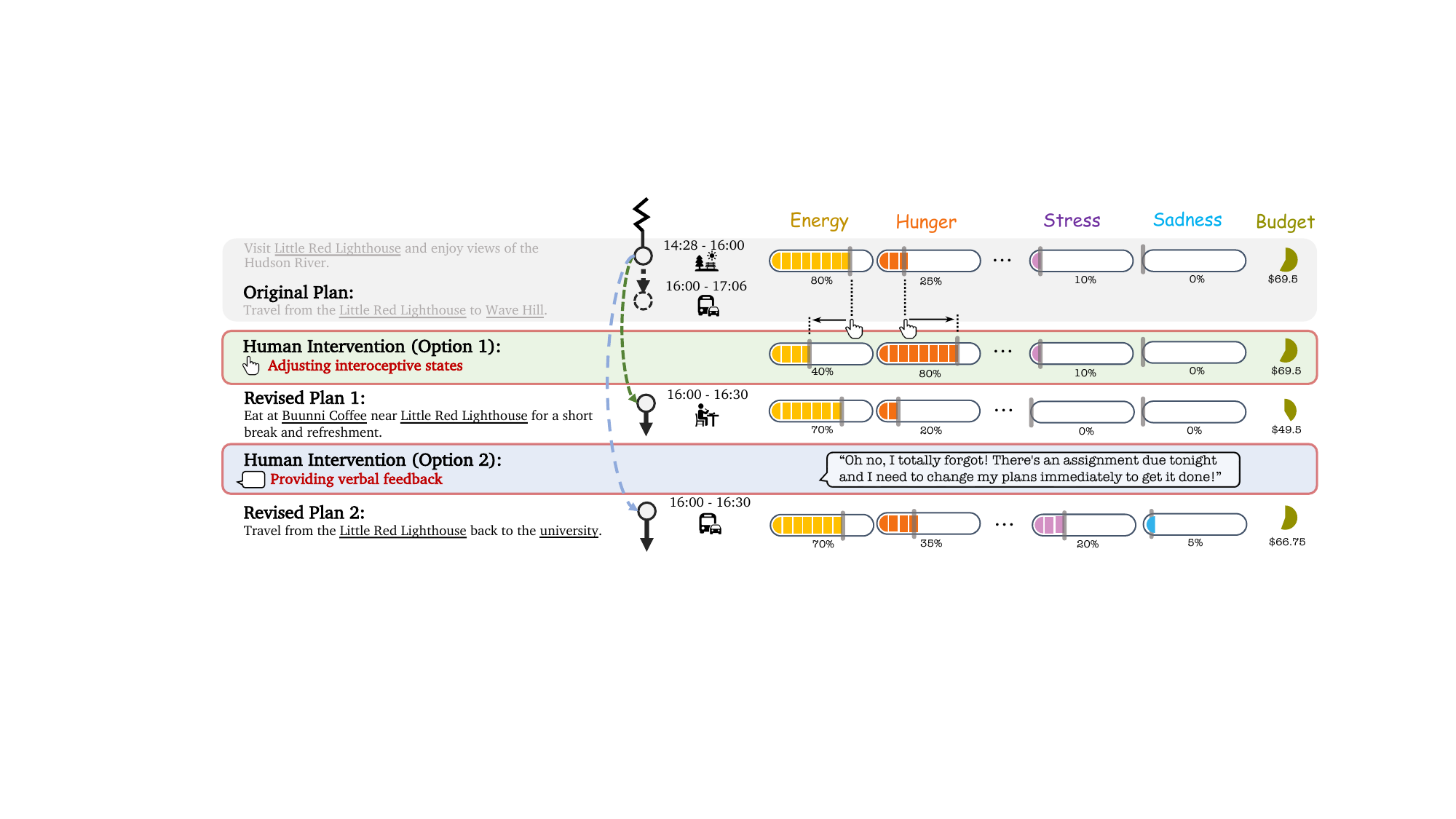}
    \vspace{-0.5cm}
    \caption{ 
    Diego adapts original plan to suit user's intervention.
    }
    \vspace{-0.1cm}
    \label{fig:agent_interactive_concierge_revise_combine}
\end{figure}

Also, as shown in \cref{fig:agent_interactive_concierge_revise_combine}, you can intervene in Diego's planning process by adjusting your interoceptive status or by providing verbal feedback.
In response, Diego promptly revises his original plan to accommodate your demands, and re-estimates your state changes after his revision.

Finally, using \virl's street views and \geoul{Map}, Diego can traverse regions of interest scouting for potential scenic viewpoints for you to visit as shown in \cref{fig:agent_photographer}. 
He uses \cvul{VQA} to rate and assess each captured view, and adds the highest-rated locations to your itinerary.

\section{System Fundamentals}
\label{sec:system}
This section introduces our system's core: a platform designed for perception-driven agents that transforms real-world cities around the world into a vast virtual playground where agents can be constructed to solve practical tasks.
At its heart, \virl is comprised of a hierarchical architecture (see \cref{fig:architecture}). 
The \textit{platform} lies at the foundation---providing the underlying components and infrastructure for agents to employ.
Higher level \textit{capabilities} of \cvtag{Perception}, \llmtag{Reasoning}, \geotag{Action}, and \colabtag{Collaboration} emerge from the platform's components.
Finally, \textit{agents} leverage these capabilities and user-defined metadata in task-specific routines to solve tasks.

\subsection{Agent Definition}
\label{sec:system_agent}
In our system, agent behavior is shaped by user-defined metadata, including a background, an intended goal, and an interoceptive state.
The \emph{background} provides the context necessary to instantiate the agent in the real world (location), and to guide its reasoning and decision-making (biography).
\emph{Intentions} outline agents' purpose within the environment.
An agent's \emph{interoceptive state} reflects its internal mental and physical status---varying over time and influencing its behavior.
This novel concept is crucial to AI agents for enhancing collaboration with humans (see \cref{sec:virl_human_agent}).

Concretely, agents are developed by writing task-specific \texttt{run()} routines that leverage the various components of our platform and the agent's metadata to solve tasks.

\subsection{Platform Components}
\label{sec:system_platform}
Next, we delve into the platform components, which provide the infrastructure to instantiate capabilities, execute agent actions, and ground agents in the real world.

\begin{figure}[h!]
    \centering
    \includegraphics[width=1\linewidth]{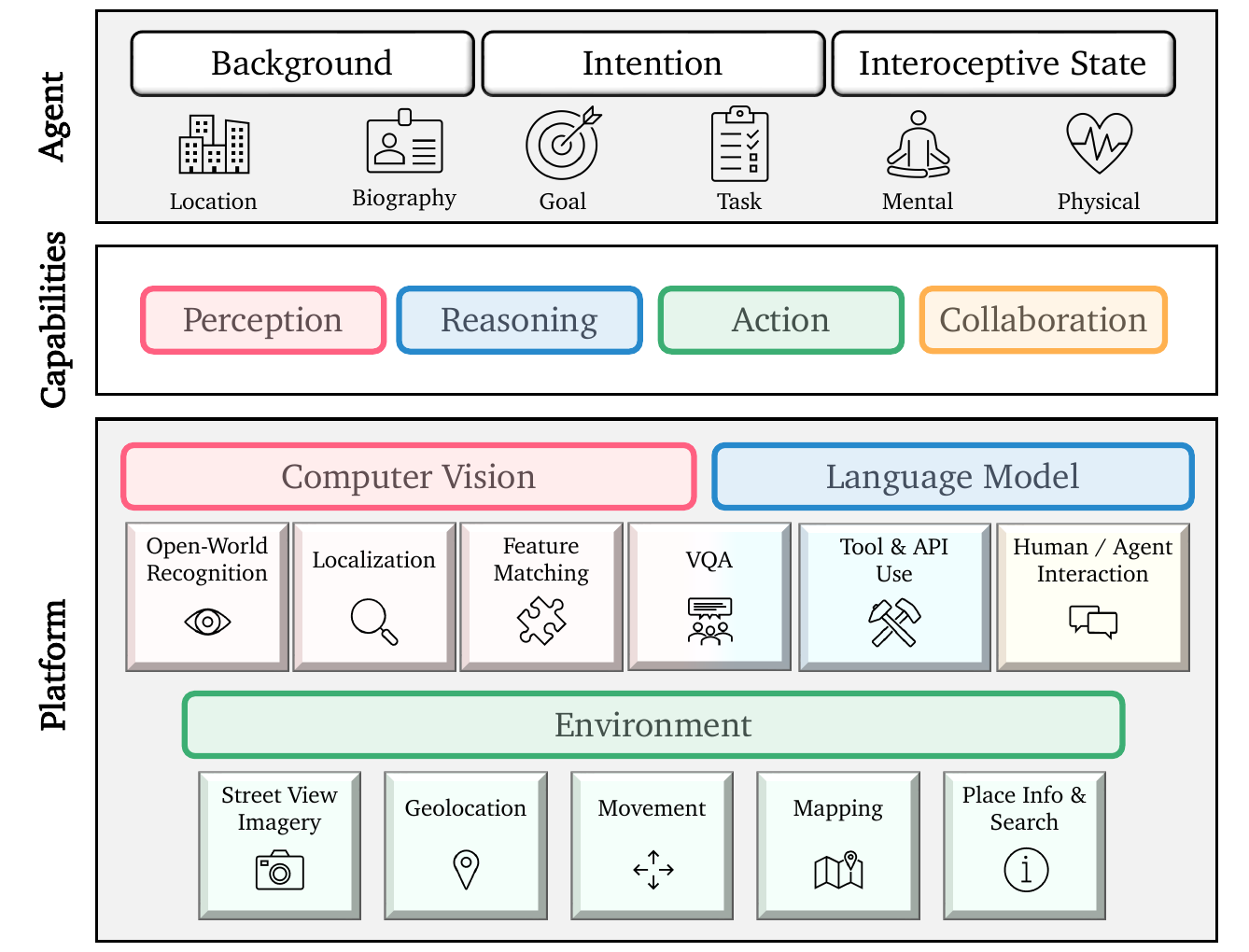}
    \vspace{-0.4cm}
    \caption{
        Hierarchical \virl architecture described in \cref{sec:system}.
    }
    \vspace{-0.3cm}
    \label{fig:architecture}
\end{figure}

\vspace{-0.5cm}
\subsubsection{{Environment (Action)}}
\vspace{-0.2cm}
\label{sec:system_platform_environment}
\geotag{Environment} components are responsible for grounding agents in the world around them: providing a navigable representation of real cities (see \cref{sec:virl_real_world}).
Geographic coordinates serve as the link between the world and our virtual representation of it.
Leveraging the Google Maps Platform (GMP)~\cite{google_map_platform}, \virl enables agents to access street view imagery, query valid movements, retrieve information about nearby locations, and plan routes.
As these coordinates and location information are bound to the real world, they also provide a natural interface with external tools that leverage geolocation---such as real estate APIs (see \cref{sec:virl_lang_driven}).
Technical designs of environment are detailed in \cref{sec:environment_detials}.

\subsubsection{{Vision (Perception)}}
\label{sec:system_platform_perception}
\cvtag{Perception} components enable
agents to process the sensory-rich data provided by the \geoul{environment}, especially street view imagery.
Pretrained localization models~\cite{li2022grounded} give agents a precise spatial understanding of their environment.
This allows RX-399 to identify and count instances of objects, and Hiro to pick out specific businesses to look up with the \geoul{GMP} (\cref{sec:virl_visual_grounding}).
While localization models allow for precise interaction with perceptive input, open-world recognition models~\cite{radford2021learning} are more general, and allow agents to detect a wider range of objects in their field of view (\eg, Tourist searches for the Apple Store).
Pretrained feature matching models~\cite{lindenberger2023lightglue}
provide an understanding of continuity across views of the same location, and
enable agents to identify \& deduplicate instances of the same object from different viewpoints (\cref{sec:virl_visual_grounding}).
Multimodal models with VQA \& Captioning capabilities~\cite{li2023blip}
bridge the perceptual world with natural language, and
are essential for integration with \llmul{reasoning} (\cref{sec:virl_visual_grounding}).

\begin{figure*}[t!]
    \centering
    \includegraphics[width=0.7\linewidth]{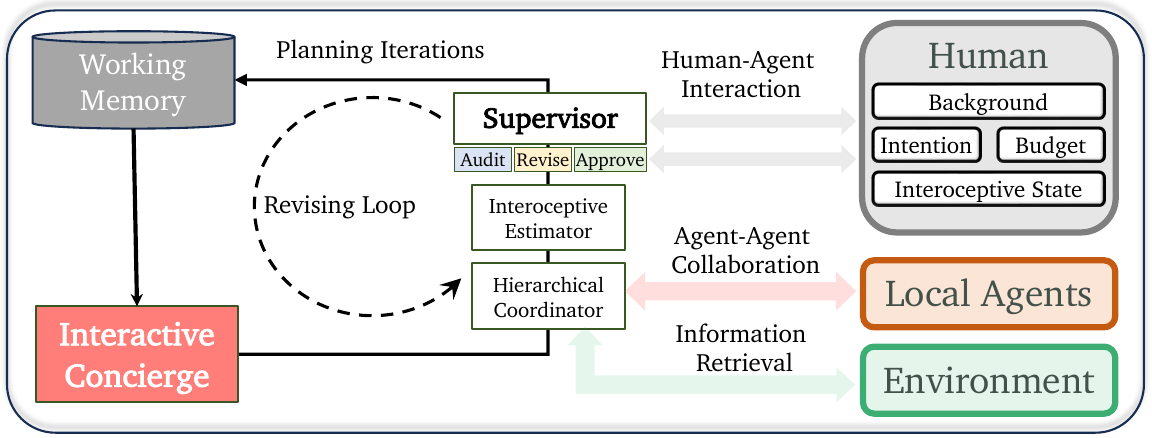}
    \vspace{-0.1cm}
    \caption{Architecture overview of interactive concierge agent Diego (\cref{sec:virl_human_agent}).
    See pipeline description in \cref{sec:system_case_study}.
    }
    \label{fig:agent_concierge_pipeline}
    \vspace{-1em}
\end{figure*}

\subsubsection{{Language (Reasoning \& Collaboration)}}
\label{sec:system_platform_reasoning}
\llmtag{Reasoning} components allow decision making based on information from \cvul{perception} and the \geoul{environment}.
LLMs such as GPT-4~\cite{openai2023gpt} and Llama~2~\cite{touvron2023llama}
interface across various APIs (\cref{sec:virl_lang_driven}), transforming environmental data and perceptual outputs into actionable insights.
They also enable
\colabtag{Collaboration}
between agents or with humans through natural language (\cref{sec:virl_collaborative})
Custom prompts facilitate this interaction (see \cref{sec:system_case_study}).

\subsection{\virl Capabilities}
\label{sec:system_capabilities}
Our platform's components
can be flexibly combined to
exhibit a vast array of capabilities.
In \cref{sec:virl_agents}, we present agents that exhibit increasingly complex behaviors, each requiring more components of the platform.
From simple combinations, like the Route Optimizer (\cref{sec:virl_real_world}),  
to more complex arrangements, 
like the Tourist (\cref{sec:virl_agent_agent}), our system showcases the versatility and potential of the \virl platform to be applied to various real-world scenarios.
Next, we perform a high-level case study of how \virl's components are combined to create our most complex agent;
in \cref{sec:system_case_study_hiro}, we delve deeper into the low-level platform details that underpin creating a \virl agent.

\subsection{High-Level System Case Study: Interactive Concierge ``Diego''}
\label{sec:system_case_study}
By studying Diego (\cref{sec:virl_human_agent}), we illustrate how our platform's components are combined to create complex agents.

Behind Diego's proficiency in developing itineraries is his iterative planning pipeline (depicted in \cref{fig:agent_concierge_pipeline}). 
The process begins with Diego creating an initial draft plan for the first activity using \llmul{GPT-4}, taking into account the user's biography, requirements, and previous activities in working memory.
This draft is then meticulously refined.
First, a \texttt{hierarchical coordination} module \geoul{retrieves} real transportation time and \colabul{asks} a recommendation agent for dining recommendations.
Subsequently, an \texttt{interoceptive estimation} module \llmul{evaluates} the effect of the proposed activity on the user's mental/physical state and budget.

The crucial final step involves a \texttt{supervisor} module, which \llmul{reviews} (``audits'') the incoming activity in light of the current user status, remaining budget, and potential interactions (exemplified in \cref{fig:agent_interactive_concierge_revise_combine}).
If the \texttt{supervisor} \llmul{deems} the plan unsuitable, it initiates revisions.
The revised plan is then looped back to the \texttt{hierarchical coordinator} and \texttt{interoceptive estimator} for reliability, followed by another review from the \texttt{supervisor} (see the revising loop in \cref{fig:agent_concierge_pipeline}).
This iterative process between the \texttt{hierarchical coordinator}, the \texttt{interoceptive estimator}, and the \texttt{supervisor} continues until the \texttt{supervisor} \llmul{approves} the activity and adds it to its \texttt{working memory}.

After finalizing an activity, Diego proceeds to plan the subsequent activity by repeating this process until the day's itinerary is complete.

\section{\virl Benchmarks}

In the previous sections, we illustrate the primary benefit of the \virl platform: seamless access to first-person street-view imagery and descriptive information about real-world cities across the globe.
This \emph{scalable} source of \emph{truly open-world} data can be harnessed to test core component models and agent capabilities.
We propose three \virl benchmarks: two evaluating vision-language models on open-world vision tasks (\cref{sec:benchmark_localization,sec:benchmark_rec_vqa}), and one evaluating end-to-end agent performance (\cref{sec:benchmark_vln}). Benchmark details are in \cref{sec:benchmark_details}.

\subsection{Automated Data and Annotation Collection}
\label{sec:benchmark_data_collection}
To allow our \virl benchmarks to scale globally, we develop an automatic data/annotation construction pipeline instead of crawling and manually annotating limited data. This allows models to be conveniently tested worldwide, provided there is access to Google Street Views~\cite{google_map_platform}.

\vspace{0.05in}
\noindent\textbf{Region Selection.}
Though our benchmark is feasible across all regions covered by the GMP, we select 14 districts across 12 cities from 6 continents to ensure coverage of a diverse data distribution while keeping inference costs affordable. The detailed locations of these regions are listed in \cref{tab:benchmark_region_list}.

\vspace{0.05in}
\noindent\textbf{Place Types.} We collect place information in each region for all 96 places types annotated by GMP\footnote{\href{https://developers.google.com/maps/documentation/places/web-service/supported\_types/\#table1}{https://developers.google.com/maps/documentation/places/web-service/supported\_types/\#table1}}. 
Our \virl place: detection, recognition and VQA benchmarks are built upon all or part of these place types.

\vspace{0.05in}
\noindent\textbf{Vision and Place Data Collection.}
Within each region, we collect geolocations with available street views, place information, and place-centric images. 

\vspace{0.05in}
\noindent\textbf{Data Cleaning.} Though scalable, automated data collection can introduce noise due to the absence of human supervision. To this end, we design three automatic data cleaning strategies: $i$) \emph{distance-based filtering} to exclude places not easily visible from any street views due to their distance; $ii$) \emph{human-review filtering} to remove ``zombie'' places with no reviews which might no longer be valid or relevant; and $iii$) \emph{CLIP-based filtering} to retain only \emph{place-centric images} with a high CLIP likelihood of being \texttt{storefronts}. 

\vspace{0.05in}
\noindent\textbf{Human Verification.} To validate the quality, we randomly sample over 10\% of data from each benchmark. We find that only about 7\% of the samples contain errors, which confirms the high quality of our data, particularly given the real-world sources.

\begin{table}[h!]
\vspace{0.2cm}
    \scalebox{0.93}{
    \begin{tabular}{lll}
    \bottomrule[1pt]
    \textbf{Continent} & \textbf{City} & \textbf{District} \\ 
    \hline
    \multirow{2}{*}{Africa} & Johannesburg & Rosebank \\ 
     & Lagos & Surulere \\ 
    \hline
    \multirow{4}{*}{Asia} & Mumbai & Khar \\ 
     & New Delhi & Lajpat Nagar \\ 
     & Hong Kong & Prince Edward \\ 
     & Tokyo & Shinjuku \\ 
    \hline
    \multirow{2}{*}{Australia} & Melbourne & CBD \\ 
     & Melbourne & SouthBank \\ 
    \hline
    \multirow{2}{*}{Europe} & Milan & Brera \\ 
     & London & Oxford St \\ 
    \hline
    \multirow{3}{*}{North America} & New York City & Chinatown, Manhattan \\ 
     & New York City & SoHo, Manhattan \\ 
     & San Francisco & Union Square \\ 
    \hline
    South America & Buenos Aires & Monserrat \\ 
    \toprule
    \end{tabular}
    }
    \caption{Region list for global \virl benchmarks.}
    \vspace{-0.2cm}
    \label{tab:benchmark_region_list}
\end{table}

\subsection{\virl Place: Detection}
\label{sec:benchmark_localization}
Every day, humans traverse cities, moving between varied places to fulfill a range of goals, 
like the Intentional Explorer agent (\cref{sec:virl_visual_grounding}).
We assess the performance of vision models on the everyday human activity of \textit{localizing places} using street view imagery and associated place data.

\vspace{0.05in}
\noindent\textbf{Setups.} We modify RX-399 (\cref{sec:virl_visual_grounding}) to traverse polygonal areas while localizing \& identifying 20 types of places. We subsample 28 polygonal areas from the 14 districts.

\vspace{0.06in}
\noindent\textbf{Benchmarked Models.}
We evaluate three prominent open-world detection models:  GroundingDINO~\cite{liu2023grounding},  GLIP~\cite{li2022grounded} and Owl-ViT~\cite{minderer2022simple}, OpenSeeD~\cite{zhang2023openseed} and Owl-ViT v2~\cite{minderer2024owlvitv2}. We also implement a straightforward baseline, CLIP (w/ GLIP proposal), which involves reclassifying the categories
of GLIP proposals with CLIP~\cite{radford2021learning}.

\vspace{0.06in}
\noindent\textbf{Evaluation.}
We evaluate the models based on localization recall, which is quantified as $\frac{N_\text{tp}}{N_\text{tp} + N_\text{fn}}$, where $N_\text{tp}$ and $N_\text{fn}$ represents the number of correctly localized places and missed places, respectively.

\vspace{0.06in}
\noindent\textbf{Matching between Object Proposals and Places.}
As mentioned in ~\cref{sec:benchmark_data_collection}, we do not annotate bounding boxes for places on each potential street view image. Such human annotation diverges from our initial motivation of providing plug-and-play and sensor-rich (\virl) benchmarks.
To assign ground truth for each object proposal in this scenario, we develop a simple matching strategy to assign object proposals from street view object detections to nearby places.

\begin{figure}[h!]
    \centering
    \includegraphics[width=0.9\linewidth]{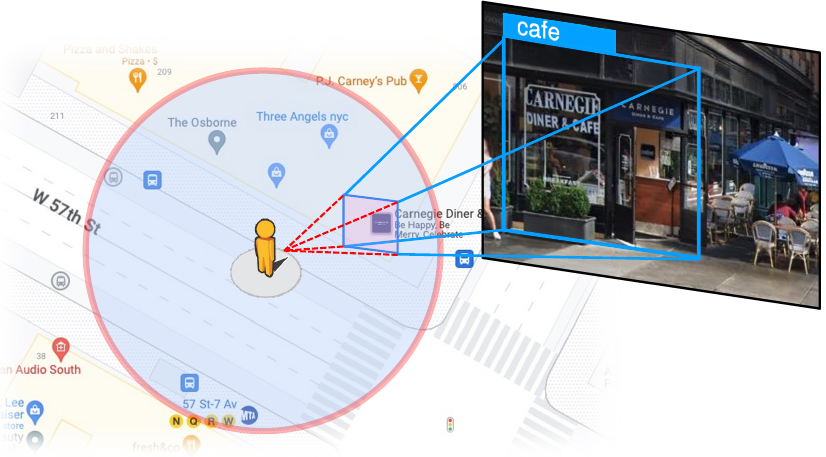}
    \vspace{-0.2cm}
    \caption{Matching between 2D object proposal and street place.}
    \label{fig:bm_localization_matching_example}
    \vspace{-0.2cm}
\end{figure}

As illustrated in \cref{fig:bm_localization_matching_example}, we first \geoul{project} the bounding box of each \cvul{object proposal} onto a frustum in the 3D space, subject to a radius.
We then determine if any \geoul{nearby places} fall within this frustum and radius. 
If any nearby place is found, the closest one is assigned as the \emph{ground truth} for the object proposal. Otherwise, the object proposal is regarded as a \emph{false positive}. 
When multiple places are inside the frustum, we consider the nearest one as the ground truth since it would likely block the others in the image.
\emph{This process is also used in Intentional Explorer agent Hiro to parse object proposals on image to place information.}

\vspace{0.05in}
\noindent\textbf{Results.}
\cref{tab:benchmark_det_res} shows that open-world detectors like GroundingDINO~\cite{liu2023grounding}, Owl-ViT~\cite{minderer2022simple} and GLIP~\cite{li2022grounded} are biased towards certain place types such as \texttt{school}, \texttt{cafe}, and \texttt{convenience store}, respectively. 
In contrast, CLIP (w/ GLIP proposal) can identify a broader spectrum of place types.
This is mainly caused by the category bias in object detection datasets with a limited vocabulary. 
Hence, even if detectors like Owl-ViT are initialized with CLIP, their vocabulary space narrows down due to fine-tuning. 
These results suggest that cascading category-agnostic object proposals to zero-shot recognizers appears promising for ``real'' open-world detection---especially for less common categories in object detection datasets.

\begin{table}[htbp]
    \centering
    \begin{small}
    \setlength\tabcolsep{1.5pt}
    \scalebox{0.68}{
        \begin{tabular}{l|cccccccccc|cc}
            \toprule
            \textbf{Place Types} & \includegraphics[width=0.3cm,height=0.3cm]{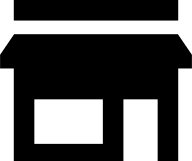} & \includegraphics[width=0.3cm,height=0.3cm]{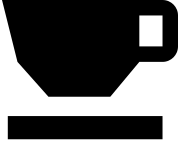} & \includegraphics[width=0.3cm,height=0.3cm]{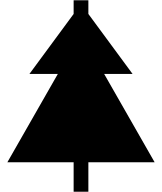} & \includegraphics[width=0.3cm,height=0.3cm]{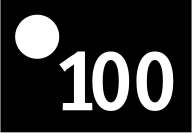} & \includegraphics[width=0.3cm,height=0.3cm]{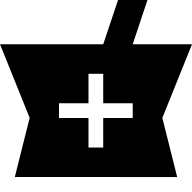} & \includegraphics[width=0.3cm,height=0.3cm]{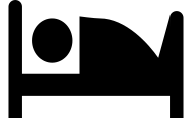} & \includegraphics[width=0.3cm,height=0.3cm]{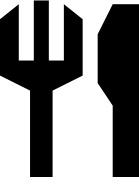} & \includegraphics[width=0.3cm,height=0.3cm]{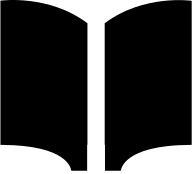} & \includegraphics[width=0.3cm,height=0.3cm]{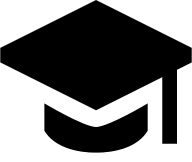} & \includegraphics[width=0.3cm,height=0.3cm]{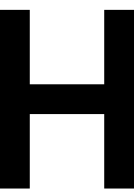} & \textbf{AR}$^{10}$ & \textbf{AR}$^{20}$ \\
            \midrule
            GroundingDINO~\cite{liu2023grounding} & 0.0 & 0.0 & 0.0 & 0.0 & 0.0 & 4.9 & 0.0 & 0.0 & \cellcolor{mygray}{100.0} & 0.0 & 11.7 & 5.8\\
            Owl-ViT~\cite{minderer2022simple} & 0.0 & \cellcolor{mygray}{61.0} & 0.0 & 0.0 & 0.0 & 2.4 & 0.3 & 0.0 & 0.0 & 0.0 & 7.1 & 7.1 \\
            GLIP~\cite{li2022grounded} & 20.0 & 0.0 & \cellcolor{mygray}{100.0} & 0.0 & 0.0 & 0.0 & 18.4 & 0.0 & 0.0 & 0.0 & 15.4 & 9.0 \\
            OpenSeeD~\cite{zhang2023openseed} & \cellcolor{mygray}{60.0} & 11.9 & 50.0 & 0.0 & 0.0 & 0.0 & \cellcolor{mygray}{20.5} & 0.0 & 0.0 & 16.7 & 17.7 & 16.7 \\
            Owl-ViT v2~\cite{minderer2024owlvitv2} & 0.0 & 0.0 & 0.0 & 0.0 & 0.0 & 0.0 & 0.0 & 0.0 & 0.0 & 0.0 & 0.0 & 3.4 \\
            CLIP~\cite{radford2021learning} (w/ GLIP proposal) & \cellcolor{mygray}{60.0} & 6.8 & 50.0 & \cellcolor{mygray}{40.0} & \cellcolor{mygray}{25.0} & \cellcolor{mygray}{29.3} & 14.7 & 0.0 & 0.0 & \cellcolor{mygray}{16.7} & \cellcolor{mygray}{26.9} & \cellcolor{mygray}{23.7} \\
            \bottomrule
            \end{tabular}
    }
    \end{small}
    \caption{
        Benchmark results on \virl Place Detection. AR$^{10}$ and AR$^{20}$ denote average recall on subsampled 10 and all 20 place categories, respectively. Full results in \cref{sec:benchmark_details_localization}.
    }
    \label{tab:benchmark_det_res}
    \vspace{-0.4cm}
\end{table}

\subsection{\virl Place: Recognition and VQA}
\vspace{-0.2cm}
\label{sec:benchmark_rec_vqa}
In contrast to the challenging \virl place detection task using street view imagery alone, in real life, humans can recognize businesses
by taking a closer, place-centric look.
We assess existing vision models in this manner on two perception tasks based on place-centric images: $i$) recognizing specific place types; $ii$) identifying human intentions via Vision Question Answering (VQA).

\begin{figure}[h!]
    \centering
    \includegraphics[width=1.0\linewidth]{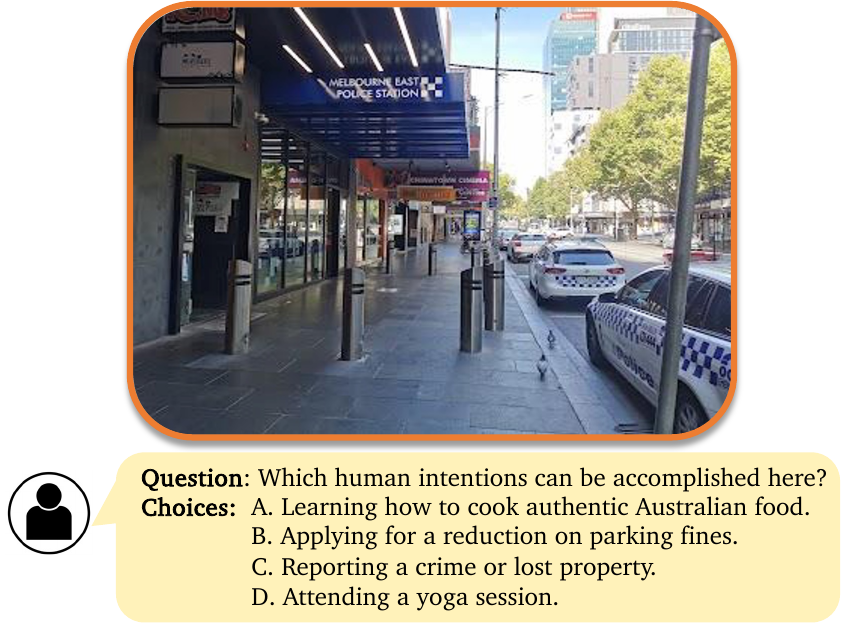}
    \vspace{-0.6cm}
    \caption{Example of \virlplace VQA process.}
    \label{fig:vqa_generate_example}
\end{figure}

\vspace{0.05in}
\noindent\textbf{Setups.}
For recognition, we assess 10 open-world recognition models on identifying a place's type (from 96 options) using place-centric images (see \cref{tab:benchmark_rec_vqa_res}). 
For place VQA, we evaluate 13 multi-modal large language models (MM-LLM) to determine viable human intentions from a four-option multiple-choice. The \virlplace VQA process is illustrated in \cref{fig:vqa_generate_example}, where the candidate and true choices are generated by GPT-4~\cite{openai2023gpt} given the place types and place names corresponding to the image.

\begin{figure}[h!]
    \centering
    \includegraphics[width=1.0\linewidth]{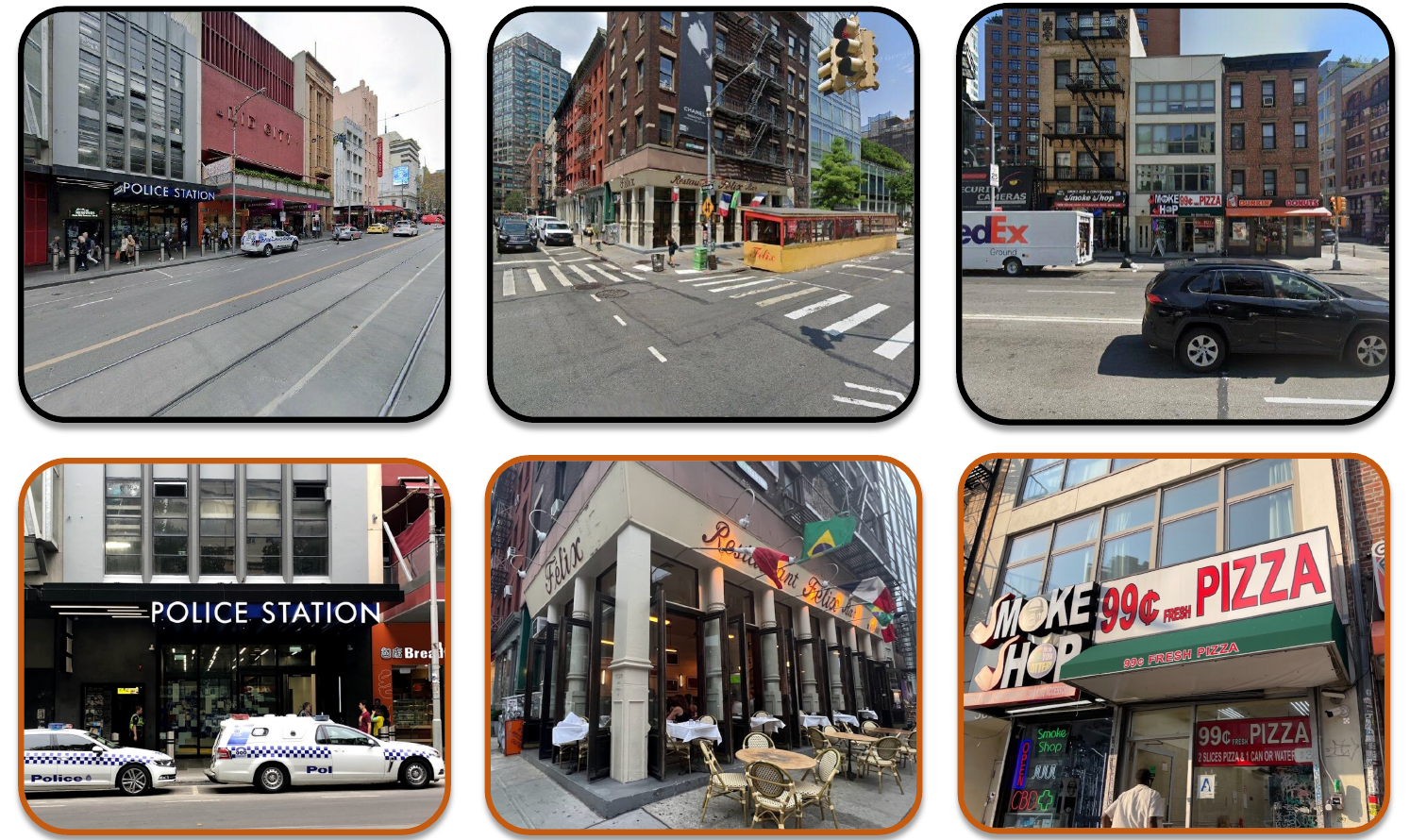}
    \caption{Top row: examples of street view imagery. Bottom row: corresponding place-centric images.}
    \label{fig:bm_place_images_vs_street_images}
    \vspace{-0.5cm}
\end{figure}

\noindent\textbf{Place-centric Images \vs Street View Images.}
In contrast to the street view imagery utilized in the \virlplace detection benchmark, the \virlplace recognition and VQA benchmarks use place-centric images. To illustrate the distinction between these image types, we present examples in \cref{fig:bm_place_images_vs_street_images}. 
The figure shows that street view images, sourced from the Google Street View database\footnote{\href{https://developers.google.com/maps/documentation/streetview/request-streetview}{https://developers.google.com/maps/documentation/streetview/request-streetview}}, are taken from the street and encompass a broader view of the surroundings, including multiple buildings and possible occluding objects/vehicles. 
In contrast, place-centric images, drawn from the Google Place database\footnote{\href{https://developers.google.com/maps/documentation/places/web-service/photos}{https://developers.google.com/maps/documentation/places/web-service/photos}}, are taken on foot and focus more closely on the specific place---providing a more concentrated view.

\vspace{0.05in}
\noindent\textbf{Evaluation.}
We adopt mean accuracy (mAcc) to evaluate both place recognition and VQA tasks.
For place VQA, we follow MMBench~\cite{liu2023mmbench} to conduct circular evaluation and GPT-assisted answer parsing.

\begin{table}[htbp]
    \centering
    \begin{small}
    \setlength\tabcolsep{10pt}
    \scalebox{0.83}{
        \begin{tabular}{llrc}
            \toprule[1pt]
            & \textbf{Model} & \textbf{\#Param} & \textbf{mAcc (\%)} \\
            \midrule
            \multicolumn{4}{c}{\textit{\textbf{\virlplace Recognition}}} \\
            \clipcell{CLIP}~\cite{radford2021learning} & \sizescell{ViT-B/32} & 151M & 18.2 \\
            \clipcell{CLIP}~\cite{radford2021learning} & \sizemcell{ViT-L/14} & 428M & 37.2 \\
            \clipcell{CLIP}~\cite{radford2021learning} & ViT-L/14@336px & 428M & \cellcolor{mygray}41.3 \\
            \openclipcell{OpenCLIP}~\cite{openclip2023} & \sizescell{ViT-B/32} & 151M & 21.2 \\
            \openclipcell{OpenCLIP}~\cite{openclip2023} & \sizemcell{ViT-L/14} & 428M & 31.0 \\
            \evaclipcell{Eva-02-CLIP}~\cite{EVA-CLIP} & ViT-B/16 & 150M & 19.5 \\
            \evaclipcell{Eva-02-CLIP}~\cite{EVA-CLIP} & \resscell{ViT-L/14} & 428M &  34.2 \\
            \evaclipcell{Eva-02-CLIP}~\cite{EVA-CLIP} & \resmcell{ViT-L/14@336px} & 428M & 40.7 \\
            \siglipcell{SigLIP}~\cite{zhai2023siglip} & ViT-B/16 & 203M & 29.5 \\
            \siglipcell{SigLIP}~\cite{zhai2023siglip} & ViT-L/16@384px & 652M & 37.3 \\
            \midrule
            \multicolumn{4}{c}{\textit{\textbf{\virlplace VQA}}}\\
            MiniGPT-4~\cite{zhu2023minigpt} & Vicuna-13B-v0 & 14B & 3.9 \\
            mPLUG-Owl~\cite{ye2023mplug} & LLaMA-7B & 7B & 5.5 \\
            Shikra~\cite{chen2023shikra} & Vicuna-7B & 7B & 10.9 \\
            BLIP-2~\cite{li2023blip} & FlanT5$_\text{XXL}$ & 12B & 69.6 \\
            InstructBLIP~\cite{dai2023instructblip} & FlanT5$_\text{XXL}$ & 12B & 68.0 \\
            LLaVA~\cite{liu2023visual} & Vicuna-13B-v1.3 & 13B & 23.5 \\
            \llavacell{LLaVA-1.5}~\cite{liu2023improvedllava} & \sizescell{Vicuna-7B-v1.5} & 7B & 60.1 \\
            \llavacell{LLaVA-1.5}~\cite{liu2023improvedllava} & \sizemcell{Vicuna-13B-v1.5} & 13B & 61.9 \\
            \llavacell{LLaVA-NeXT}~\cite{liu2024llavanext} & Vicuna-7B-v1.5 & 7B & 65.9 \\
            Mini-Gemini~\cite{li2024mini} & Vicuna-7B-v1.5 & 7B & 44.1 \\
            Intern-VL-1.5~\cite{chen2024internvl1.5} & InternLM2-20B & 26B & \cellcolor{mygray}{77.6} \\
            GPT-4V~\cite{openai2023gpt} & UNK. & UNK. & 77.1 \\
            Qwen-VL-max~\cite{bai2023qwen} & UNK. & UNK. & 74.3 \\
            \bottomrule
        \end{tabular}
    }
    \end{small}
    \caption{Benchmark results on \virlplace recognition and \virlplace VQA. \colorbox{resscolor}{Green} indicates increased resolution models, while \colorbox{sizescolor}{Blue} denotes model parameter scaling.}
    \label{tab:benchmark_rec_vqa_res}
\end{table}

\vspace{0.05in}
\noindent\textbf{Results.}
\cref{tab:benchmark_rec_vqa_res} shows that CLIP (L/14@336px) outperforms even the biggest version of Eva-02-CLIP and SigLIP in the \virl recognition task, highlighting the high-quality data used to train CLIP~\cite{radford2021learning}.
The bottom of the table shows that LLaVA-NeXT (7B) outperforms its predecessors LLaVA-1.5 and 1.0, but still has over 8\% gap to InternVL-1.5 with 26B parameters. Closed-source MLLMs GPT-4V and Qwen-VL-Max yield outstanding performance compared to most open-sourced models.
We note that even these top-performing MLLMs (\eg GPT-4V and Qwen-VL-Max) still suffer from inconsistent issues during the circular evaluation (see \cref{tab:vqa_consistency_analysis}).
Moreover, vision models perform better on place VQA over place-type recognition, suggesting direct prompts about human intention could be more effective for intention-driven tasks. We provide more analysis in \cref{sec:benchmark_details_rec_vqa}.

\begin{figure}[t]
    \centering
    \includegraphics[width=1\linewidth]{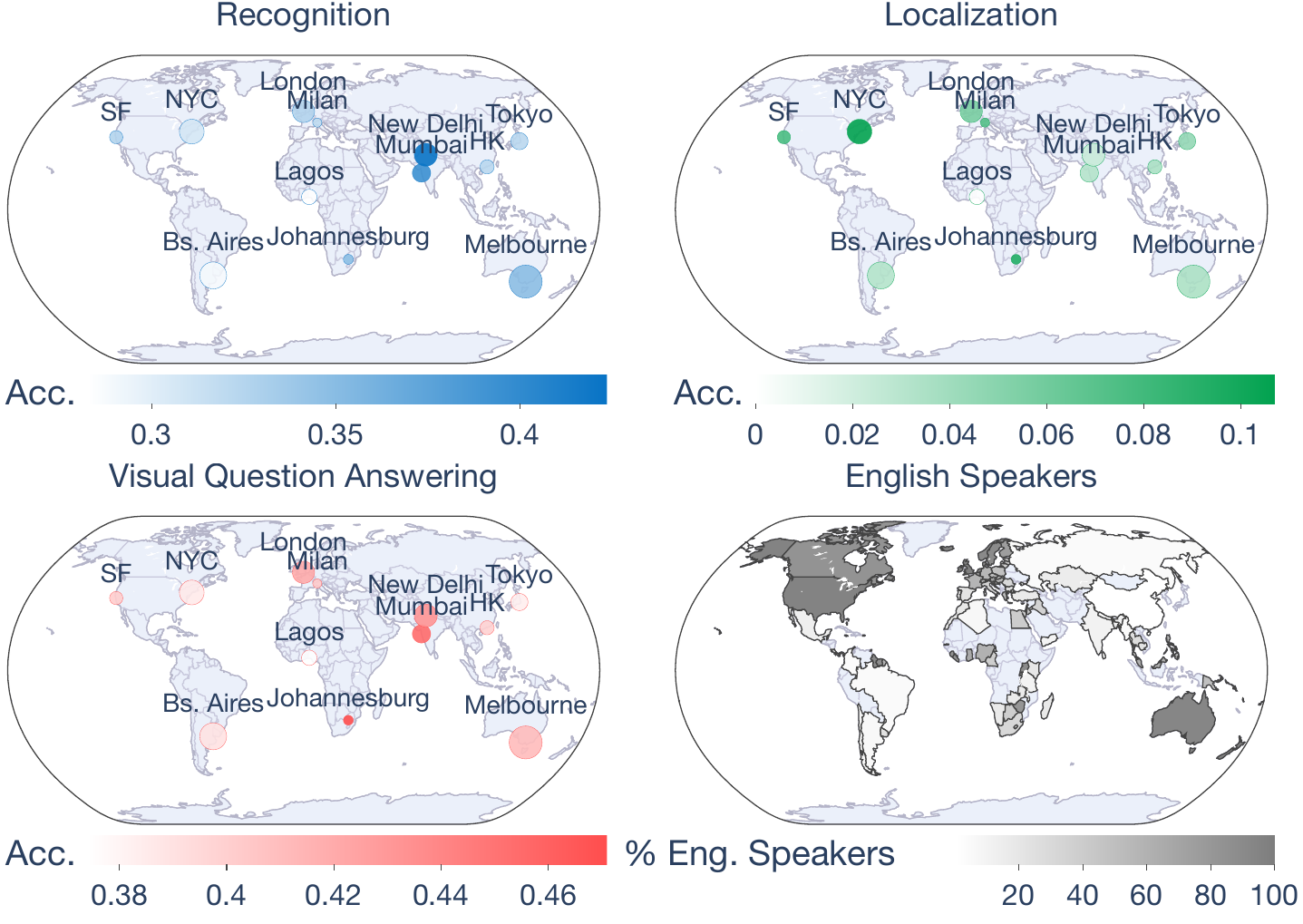}
    \vspace{-0.6cm}
    \caption{
        City-level visualization of \virl benchmark results.
    }
    \label{fig:bm_city_analysis}
    \vspace{-0.5cm}
\end{figure}

\subsection{\virl Vision-Language Navigation}
\label{sec:benchmark_vln}
As discussed in \cref{sec:virl_visual_grounding}, Intentional Explorer and Tourist agents require coordination between vision models and language models to accomplish vision-language tasks. 
To investigate the effect of various models on end-to-end agent performance, we develop an embodied task that jointly tests vision and language models: Vision-Language Navigation (VLN).
In VLN, agents navigate to a desired destination by following textual directions using only raw street views.

\vspace{0.05in}
\noindent\textbf{Setup.} 
We adopt the Tourist implementation from \cref{sec:virl_collaborative} and swap its recognition component with the various benchmarked models. These models are used to identify visual landmarks during navigation. Subsequently, 
GPT-4~\cite{openai2023gpt} predicts the next action according to the recognition results. 
Navigation instructions are generated using the Local agent.
Recent work VELMA~\cite{schumann2023velma} attempts to enhance VLN by leveraging LLMs on existing datasets~\cite{chen2019touchdown,schumann2020generating}.
In contrast, our \virl VLN benchmark evaluates vision models and their coordination with language models across a global data scale. See more details in \cref{sec:benchmark_details_vln}.

\vspace{0.05in}
\noindent\textbf{Benchmarked methods.} Four approaches are evaluated to recognize landmarks during navigation: 
($i$) Oracle that searches nearby landmarks with GMP~\cite{google_map_platform};
($ii$) Zero-shot recognizers CLIP~\cite{radford2021learning} \& EVA-CLIP~\cite{EVA-CLIP};
($iii$) Multi-modal LLM LLaVA-1.5~\cite{liu2023improvedllava};
($iv$) An OCR model~\cite{du2009pp} to extract text in street views followed by GPT answer parsing.
Implementation details are provided in \cref{sec:benchmark_details_vln}.

\vspace{0.05in}
\noindent\textbf{Evaluation.} We primarily measure navigation success rate (\textit{Success}), defining success as the navigator stopping within 25 meters of the destination. In addition, as navigation success is mainly influenced by the agent's actions at key positions (\ie, start positions, intersections and stop positions), we also evaluate the arrival ratio (\emph{Arr}) and reaction accuracy (\emph{Reac}) for each route. \emph{Arr} denotes the percentage of key positions reached, while \emph{Reac} measures the accuracy of the agent's action predictions at these key positions. To save GPT-4 resources, we mainly compare vision modules on a 10\% mini-set comprising 18 routes from 9 regions. 
See \cref{sec:benchmark_details_vln} for full-set results with CLIP and Oracle.

\begin{table}[t]
\vspace{0.2cm}
\centering
\begin{small}
\setlength\tabcolsep{3pt}
\scalebox{0.8}{
    \begin{tabular}{lcccccc}
        \toprule
        \multirow{2}{*}{\textbf{Method}} & \multicolumn{1}{c}{\textbf{}} & \multicolumn{1}{c}{\textbf{Start}} & \multicolumn{2}{c}{\textbf{Intersection}} & \multicolumn{2}{c}{\textbf{Stop}}\\
        \cmidrule{3-7}
        & \textbf{Success} & \textbf{Reac} & \textbf{Arr} & \textbf{Reac} & \textbf{Arr} & \textbf{Reac}\\
        \midrule
\cellcolor{mygray}Oracle (No Vision) & \cellcolor{mygray}1.0 & \cellcolor{mygray}1.0 & \cellcolor{mygray}1.0 & \cellcolor{mygray}1.0 & \cellcolor{mygray}1.0 & \cellcolor{mygray}1.0\\
\midrule
        CLIP (B/32)~\cite{radford2021learning} & 0.22 &  \cellcolor{mygray}1.0 & \cellcolor{mygray}0.86 & 0.84 & \cellcolor{mygray}0.83 & 0.22 \\
         CLIP (L/14@336px)~\cite{radford2021learning} & \cellcolor{mygray}0.44 &  0.83 & 0.73 & 0.94 & 0.67 &  \cellcolor{mygray}0.44 \\
        EVA-02-CLIP (BigE/14-plus)~\cite{EVA-CLIP} & 0.39 & 0.89 & 0.77 & 0.94 & 0.72 & 0.39 \\
        EVA-02-CLIP (L/14@336px)~\cite{EVA-CLIP} & 0.22 & \cellcolor{mygray}1.0 & 0.82 & 0.83 & 0.78& 0.22 \\
        \midrule
        \colorbox{resscolor}{LLaVA-1.5-13B}~\cite{liu2023improvedllava} & 0.11 & 0.61 & 0.55 & \cellcolor{mygray}1.0 & 0.56 & 0.11 \\
        \colorbox{sizescolor}{PP-OCR}~\cite{du2009pp} (+ GPT3.5) & 0.28 & \cellcolor{mygray}0.89 & 0.73 & 0.94 & 0.72 & 0.28 \\
        
        \bottomrule

    \end{tabular}}
\end{small}
\vspace{-0.2cm}
\caption{Results on \virl VLN-mini. We test various CLIP-based models, \colorbox{resscolor}{MM LLM}, and \colorbox{sizescolor}{OCR} model with GPT postprocessing.}
\vspace{-0.5cm}
\label{tab:vln_results_miniset}
\end{table}

\vspace{0.05in}
\noindent\textbf{Results.}
Table~\ref{tab:vln_results_miniset} shows that, with \geoul{oracle landmark information}, powerful LLMs can impressively comprehend navigation instructions and thus make accurate decisions.
However, when relying on vision models to fetch landmark information from street views, the success rate drops dramatically---suggesting that the perception of vision models is noisy and misguides LLMs' decision-making. 
Among these recognizers, larger variants of CLIP~\cite{radford2021learning} and EVA-02-CLIP~\cite{EVA-CLIP} perform better, highlighting the benefits of model scaling.
LLaVA-1.5~\cite{liu2023improvedllava} shows inferior performance with CLIP (L/14@336px) as its vision encoder, possibly due to the alignment tax~\cite{openai2023gpt} introduced during instruction tuning. 
Further, PP-OCR~\cite{du2009pp} (+ GPT-3.5) achieves a 28\% success rate, signifying that OCR is crucial for visual landmark recognition.

\subsection{Geographic Diversity}
\label{sec:geographic_diversity}
Spanning 12 cities across the globe, our \virl benchmarks provide an opportunity to analyze the inherent model biases across different regions. As depicted in \cref{fig:bm_city_analysis}, vision models demonstrate subpar performance on all three benchmark tasks in Lagos, Tokyo, Hong Kong, and Buenos Aires.
Vision models might struggle in Lagos due to its non-traditional street views relative to more developed cities (see street views in \cref{fig:teaser}). For cities like Tokyo, Hong Kong, and Buenos Aires, an intriguing observation is their primary use of non-English languages in street views, as shown in \cref{fig:bm_city_analysis} bottom right
\footnote{Source: \href{https://en.wikipedia.org/wiki/List\_of\_countries\_by\_English-speaking\_population}{https://en.wikipedia.org/wiki/List\_of\_countries\_by\_English-speaking\_population}}
and \cref{fig:teaser}.
This suggests that existing vision models may face challenges
when deployed in non-English-dominant countries.
\section{Discussion: Ethics \& Privacy}
Our platform serves as a tool for AI development and as a crucible for ethical discourse and preparation. 
As AI is inevitably being integrated into society---\eg, via augmented reality wearables, spatial computing platforms, or mobile robots navigating city streets---it is imperative to confront and discuss ethical and privacy concerns now.
Unlike these impending \emph{real-time} systems, the data accessed by \virl is ``stale'' and preprocessed---providing a controlled environment to study these concerns.

Notably, \virl exclusively utilizes preexisting, readily available APIs; it does not capture or make available any previously inaccessible data. 
Our primary source of street-view imagery, Google Maps~\cite{google_map_platform}, is subject to major privacy-protection measures, including blurring faces and license plates~\cite{frome2009large}.
Moreover,
\virl complies with the Google Maps Platform license\footnote{\url{https://cloud.google.com/maps-platform/terms}},
similarly to notable existing works that also leverage Google's street views~\cite{zamir2014image, chen2019touchdown}.

We believe \virl is an invaluable tool for researching bias.
As discussed in \cref{sec:geographic_diversity},
\virl's \emph{global scale} provides a lens to study linguistic, cultural, and other geographic biases inherent in models.
By using \virl to study such questions, we aim to preemptively tackle the ethical dilemmas that will arise with deploying real-time systems rather than being blindsided by them.
We hope our work helps spur proactive discussion of future challenges throughout the community.

\section{Conclusion}
In this work, we introduce \virl, an open-source platform designed to bridge the sensory gap between the digital and physical worlds, enabling AI agents to interact with the real world in a virtual yet realistic environment.
Through \virl, agents can develop rich sensory grounding and perception, utilizing real geospatial data and street-view imagery.
We demonstrate the platform's versatility by creating diverse exemplar agents and developing benchmarks measuring the performance of foundational language and vision models on open-world visual data from across the globe.

This platform opens new avenues for advancing AI capabilities in perception, decision-making, and real-world data interaction.
As spatial computing and robotic systems become increasingly prevalent, the demand for and possibilities of AI agents will only grow.
From personal assistants to practical applications like urban planning to life-changing tools for the visually impaired, we hope \virl helps usher in a new era of perceptually grounded agents.

\newpage
{
    \small
    \bibliographystyle{ieeenat_fullname}
    \bibliography{main}
}

\appendix
\clearpage

\section{Appendix Outline}
\label{sec:appendix}
In these supplementary materials, we provide additional details for our \virl platform, including:
\begin{itemize}
\item Designs behind \virl Agents (\cref{sec:agent_detials}); 
\item Technical details and challenges in the \virl environment (\cref{sec:environment_detials}). 
\item A low-level case study of Intentional Explorer agent Hiro, delving into implementation details of our system such as LLM prompts (\cref{sec:system_case_study_hiro});
\item More detailed setups and results for our \virl benchmarks (\cref{sec:benchmark_details}).
\end{itemize}

\section{Technical Details of \virl Agents}
\label{sec:agent_detials}
In \cref{sec:virl_agents}, our discussion mainly focuses on the innovative capabilities and behaviors of \virl agents empowered by our platform. 
We avoid in-depth discussions about technical details in the main paper due to the concern of readability.
In this section, we go through our main technical designs for each agent. More comprehensive technical implementations are available in our released code.

\subsection{Peng: Route Optimizer}
\label{sec:peng_details}
Peng is designed to showcase the utilization of \geoul{real geographic coordinates} within our platform. By processing a sequence of real addresses, Peng calculates the \textit{shortest path} for traversing them using \geoul{various modes of transportation}, such as walking, driving, and bicycling, among others. This capability is powered by the mapping module described in \cref{sec:mapping_details}. After that, Peng proceeds to navigate through the destinations along the predetermined path, employing the point navigation procedure outlined in \cref{sec:navigators_detail}.

\subsection{Aria: Place Recommender}
\label{sec:aria_details}
Aria leverages the \geoul{realistic place information} provided by our Place Info \& Search module (see \cref{sec:place_details}) to 
enhance \llmul{LLMs' reasoning} capability in the geographic aspect.
Specifically, Aria \llmul{evaluates} Peng's intention to determine the suitable type of place and \geoul{searches} all possible places in the vicinity. 
For each searched place, Aria \llmul{considers} its reviews and user ratings from Google to \llmul{summarize} a place overview.  
Subsequently, we customize prompts for Aria to amalgamate Peng's biography, intentions, and the summarized place overviews to \llmul{rate} each place between 0 and 10, accompanied by justifications.

Without such technical designs, LLMs could recommend some places that are either too distant or permanently closed. This issue arises because LLMs struggle to accurately understand geospatial relationships and often depend on outdated databases.

\subsection{Vivek: Estate Agent}
\label{sec:vivek_details}
The process employed by Vivek is similar to that of Aria, as both are designed to recommend places. However, Vivek showcases the versatility of our \virl platform by demonstrating how it can seamlessly integrate a wide range of realistic information beyond the Google Maps Platform, with a standardized definition of geographic coordinates. This capability enables the creation of even more sophisticated and intriguing agents.

\subsection{RX-399: Urban Assistance Robot}
\label{sec:rx399_details}
Different from previous example agents, RX-399 introduces visual perception capabilities such as \cvul{open-world detection} and \cvul{feature matching}. There are two fundamental systems inside it -- navigation and perception. In terms of navigation, RX-399 can automatically navigate from the current position to the pre-defined destination step by step. This navigation process is elaborated in \cref{sec:navigators_detail}, and thus, will not be extensively discussed here.

When it comes to its perception system, RX-399 is designed to simulate human visual perception by capturing street views within a 90-degree 
horizontal angle to both its left and right. For each captured view, an open-world detection process is conducted. 
Leveraging the \geoul{interactive capabilities} of our environment, we further propose an \cvul{\textit{active detection}} strategy to dynamically \geoul{adjust the agent's ego-pose and focal length} according to the scale and position of potential objects. This significantly improves its performance as illustrated in ~\cref{tab:rx-399_detect_results}. In the future, more advanced approaches such as visual search~\cite{wu2023vstar} could also be considered.
In the subsequent de-duplication procedure, which aims to avoid double-counting objects across different viewpoints, we have tried a few strategies including measuring with multi-view geometry, object tracking, and feature matching. We choose feature matching because of its accuracy and efficiency.

\begin{table}[h!]
\begin{center}
    \begin{small}
        \setlength\tabcolsep{8pt}
        \scalebox{1.0}{
        \begin{tabular}{c|cc}
            {City}  & Hong Kong & New York City \\
            \hline
            w/ active detection  & \textbf{0.63} / \textbf{0.83} & \textbf{0.71} / \textbf{1.00} \\
            w/o active detection & 0.10 / 0.33 & 0.30 / 0.60 \\
        \end{tabular}}
    \end{small}
    \vspace{-0.2cm}
    \caption{RX-399 detection performance with or without active detection manner. Metrics are accuracy / recall.}
    \label{tab:rx-399_detect_results}
    \vspace{-0.4cm}
\end{center}
\end{table}

\subsection{Imani: Urban Planner}
\label{sec:imani_details}
The visual perception system of Imani mirrors that of RX-399. The primary distinction between them lies in their navigation systems. 
Imani possesses the capability to plan a navigation route in the given polygonal region, enabling RX-399 to traverse that region. This functionality is named ``region navigation'' and elaborated in \cref{sec:navigators_detail}.
Additionally, within the Imani agent, we develop a heatmap visualization tool to visualize and verify the data collected by RX-399 (see \cref{fig:agent_urban_plan}).    

\subsection{Hiro: Intentional Explorer}
\label{sec:hiro_details}
Hiro is a representative agent equipped with geographical, perceptual, and reasoning abilities, to address a daily human task: 
randomly exploring to find a suitable restaurant. 
In this regard, we have dedicated a separate section to offer an in-depth case study, including the detailed methodology and prompts in \cref{sec:system_case_study_hiro}.

\subsection{Ling: Tourist}
\label{sec:ling_details}
Our vision language navigation pipeline of Ling is similar to \cite{schumann2023velma}, leveraging \cvul{vision models}, the \geoul{map}, and \llmul{LLMs}. 
At each position, we start by capturing eight street views around the agent, corresponding to \texttt{front}, \texttt{left front}, \texttt{left}, \texttt{left behind}, \texttt{behind}, \texttt{right behind}, \texttt{right} and \texttt{right front}. 
Vision models use these street views to \cvul{identify} landmarks mentioned in route descriptions, which are then verbalized as \emph{landmark observations}.
Also, intersection information is retrieved from the \geoul{mover} to formulate an \emph{intersection observation}. 
LLMs play a crucial role in \llmul{processing} landmark \& intersection observations along with the agent's previous working history to \llmul{determine} the next action.
After each action, current observations and actions are stored into the agent's working history. 
This auto-regressive process continues until the agent decides to \texttt{stop}.

\subsection{Local Agent}
\label{sec:local_details}
The primary mission of the Local agent is to generate human-like and easily followable navigation instructions on a global scale (refer to \ref{sec:virl_agent_agent}). This task is known as navigation instruction generation~\cite{schumann2020generating}. Contrary to most existing research, which depends on human-annotated data for limited geographic areas, our ``Local'' agent automatically \geoul{selects suitable landmarks} taking account into real-world places and \llmul{generates human-like route descriptions} using LLMs across the globe. Remarkably, it achieves this without the need for any training data, relying solely on our tailored prompts and a few in-context examples. 
The effectiveness of its generated instructions has been verified through collaboration with ''Ling''. To the best of our knowledge, this is a first in the field. There are massive technical details on selecting easily noticeable landmarks and prompt engineering, which are available in our released code. 

\subsection{Diego: Interactive Concierge}
In \cref{sec:system_case_study}, we have already presented the technical designs of Diego's itinerary. Here, we detail how Diego can find scenic locations as shown in \cref{fig:agent_photographer}. For any given destination, such as ``Fort Tryon Park'', Diego will sample a rectangle region around it and \geoul{traverse} all navigable positions within it. At each position, he will \geoul{capture a photograph} (\ie street view imagery) using pre-defined headings, pitches, and FOVs. Each photograph will then be evaluated using \cvul{GPT-4(V)}~\cite{openai2023gpt}, where it receives a rating between 0 and 10 along with \llmul{explanatory reasons}.

\section{Technical Details of Environment}
\label{sec:environment_detials}
In \cref{sec:system_platform_environment}, we provide an overview of our system's environment, which grounds agents in real life. Here, we delve into the technical designs beyond mere leveraging Google Map Platform system calls. Concrete implementations can be found in our open-sourced code.

\subsection{Geolocation \& Street View Imagery}
\label{sec:env_geolocation_streetivew}
At the core of \virl lies its innovative use of sensor-rich environment, including street view imagery and geolocations. They enable agents to gather surrounding place and vision information.

\vspace{0.2cm}
\noindent\textbf{Geolocation.} Agents in the \virl platform inhabit virtual representations of real cities around the globe. At the core of this representation are geographic coordinates (\ie geolocation) corresponding to points on the Earth's surface. The initial geolocation of 
each agent is specified by its ``Location'' configuration, as illustrated in \cref{fig:architecture}. 
Whenever agents require access to surrounding information (\eg street views, places or maps), geolocation serves as a crucial parameter for querying the related Google Map APIs.

\vspace{0.2cm}
\noindent\textbf{Street view imagery.} Google Map Platform specifies each street view imagery with multiple key parameters: geolocation, heading (the horizontal angle ranging from 0$^\circ$ to 360$^\circ$), pitch (a vertical angle ranging from -90$^\circ$ to 90$^\circ$), and Field of View (FOV, ranging from $20 \sim 120$).
It's noteworthy that adjusting the FOV here is similar to changing the camera's focal length, rather than simply zooming in on an image, which ensures that image resolution remains high, even as the FOV decreases to a low value.
By modifying the heading, pitch, and FOV, we can simulate the human sensory process of adjusting one's pose and concentrating on a specific area. 

\vspace{0.2cm}
\noindent\textbf{Alignment between street view imagery and geolocation.} 
Within our sensor-rich platform, a fundamental challenge is to ensure agents are positioned at geolocations where street view imagery is available.
{To address this issue, we design a custom operation named ``\textit{\textbf{relocate}}''.
Specifically, when an agent is initialized at a location lacking street view imagery, the ``relocate}'' operation will identify and transition the agent to the nearest viable geolocation where street view data is available.
Notice that, this operation is indispensable to our platform, as the positions with available street views are relatively sparse in comparison to the vast continuous space of all possible coordinates.

\subsection{Movement}
\label{sec:env_movement}
Enabling agents to move along city streets is a core functionality of our platform, allowing interaction between agents and the real world. Whenever an agent needs to move, this module powers all related processes, from route planning and direction selection to the continuous update of the agent's geolocation during its moving.
Since Google Maps Platform does not provide APIs to access nearby navigable positions and directions, the design of this movement module is a significant technical challenge and a substantial contribution from our team. 
We discuss its low-level implementations in \cref{sec:mover_detail} and the enabled high-level actions in \cref{sec:navigators_detail}.

\subsubsection{Mover}
\label{sec:mover_detail}
\noindent\textbf{Move by controlling the web interface.} A straightforward solution is to let the agent control the web front-end Google Street View to select moving directions and move. Nevertheless, there are three key challenges for this solution:

($i$) \textbf{\textit{How can Python-implemented agents control the movement via the interaction to the webpage?}} We use a Python package Selenium\footnote{\url{https://www.selenium.dev/}} to locate web elements responsible for movement. After determining a movement direction, the agent uses Selenium to simulate a click action on the web element corresponding to the chosen direction.

($ii$) \textbf{\textit{How can the agent acquire the necessary information to decide moving direction?}} Although agents can access all potential movement directions from web elements, they cannot identify these directions without prior knowledge of what each represents. We find that the ``transform'' attribute in the web element corresponding to each direction can be leveraged to calculate their represented heading angles. 
The heading angle also allows us to collect street view imagery for each movement direction. 
Agent's movement decision-making is then based on these heading angles and the visual data from street view imagery.

($iii$) \textbf{\textit{How to track the agent's geolocation along its movement?}} 
To accomplish this, we customize a webpage element to display the geolocation of the current street view panorama. As the agents move and trigger updates to the street view panorama, this customized element concurrently refreshes to reflect the new geolocation. By using Selenium, we can then extract this updated geolocation data, enabling continuous tracking of the agent's geolocation changes.

\vspace{0.2cm}
\noindent\textbf{Move by grid-based relocating.} In our test of the above web-based mover, a critical limitation emerged: the web-embedded Google Street View panoramas display only a subset of navigable directions. This constraint significantly restricts our agents' mobility, often preventing them from successfully navigating to their intended destinations due to the incomplete coverage of potential routes.

To overcome this obstacle, we develop an alternative method: a grid-based relocating mover. This approach involves performing a grid search for geolocations in the vicinity of the agent and employing the ``relocate'' operation to sift through these locations, identifying those that are navigable. While this method offers a more comprehensive view of navigable positions, it is markedly more time-consuming than the web-based approach due to the extensive number of Google Maps API calls required.

In our practical applications, we design a heuristic strategy that combines web-based controlling and grid-based relocation. This hybrid approach aims to balance the trade-offs between the speed and the completeness of navigable position data, optimizing our agents' capabilities and efficiency in real-world scenarios.

\subsubsection{Navigator}
\label{sec:navigators_detail}
Here, we introduce the high-level action of agents powered by the mover -- navigation. Unlike the mover,  which concentrates on enabling agent mobility in the environment, the focus here shifts to determining the direction of movement. In our platform, we group different navigators according to their usages into four types:

($i$) \textbf{Point navigator} is designed to tackle navigation tasks that clearly define single or multiple destinations (represented in addresses or geolocations). This navigator employs the route planning function described in \cref{sec:mapping_details} to obtain a series of key positions for navigation. At each location, the agent utilizes a greedy algorithm to select the most optimal direction towards the next key position that has not yet been reached.
Exemplary agents, such as ``Peng'', ``RX-399'' and ``Local'', use this type of navigator in their implementation.

($ii$) \textbf{Region navigator} is tailored for agents like ``Imani'' and ``Diego'', who need to traverse every position within a polygonal region. This navigator first employs a grid search combined with our ``relocate'' operation to identify all navigable positions within the specified region. Subsequently, it adopts a heuristic algorithm designed to solve the traveling salesman problem, planning an efficient order for visiting these positions. The agents' task is to simply follow this predetermined route, visiting each navigable position in the planned order.

($iii$) \textbf{Vision-language navigator} is specifically developed for the tourist agent ``Ling'', as well as for tasks within the \virl vision-language navigation benchmark. Its primary function is to guide the agent in selecting a proper direction based on navigation instructions. The detailed pipeline is presented in \cref{sec:ling_details}.

($iv$) \textbf{Intention navigator} is utilized in intentional explorer agent ''Hiro`` to select the most suitable direction that aligns with the agent's specific intentions. The detailed methodology and prompt are detailed in \cref{sec:system_case_study_hiro_road_selection}.

\subsection{Mapping}
\label{sec:mapping_details}
The mapping module in our environment is designed to equip agents with functionalities such as route planning and transportation time estimation. It mainly utilizes the  ``Directions API''\footnote{\url{https://developers.google.com/maps/documentation/directions}} from the Google Map Platform to facilitate these capabilities. 
Given the complex nature of this API's interface, our principal focus has been on parsing its output and adapting it into various user-friendly interfaces for agents.

\subsection{Place Info \& Search}
\label{sec:place_details}
Place Info \& Search module hosts another important information source in our platform beyond the visual street view imagery, enabling agents to interact with real-world ``places''. It provides various attributes of places, including type, name, location, imagery, reviews, etc. 
In this module, our technical efforts are primarily focused on understanding, comparing, and integrating the most suitable functions from the vast array of Google Maps Platform APIs related to place information and nearby place searches. 
Additionally, we devise some post-processing strategies to identify and eliminate invalid or conflicting data sources from the Google Maps Platform.

Another essential capability enabled by this module is to associate object proposals in street view imagery and their corresponding places in the real city. This function is vital to enhance the reality of our platform by connecting street view and geolocation. It also powers the ``Hiro'' agent and the evaluation of the \virlplace detection benchmark. The implementation is detailed in \cref{sec:benchmark_localization}.

\section{Low-Level System Case Study:\\Intentional Explorer ``Hiro''}
\label{sec:system_case_study_hiro}
This section delves deeper into the low-level implementation details of the Intentional Explorer agent ``Hiro'' (\cref{sec:virl_visual_grounding}), focusing on the prompts utilized to interact with various parts of our system. Concretely, we present the prompts in four subparts: \emph{identifying a type of place to search using the user-defined intention} (\cref{sec:system_case_study_hiro_intention_to_place_type}),
\emph{selecting appropriate roads} (\cref{sec:system_case_study_hiro_road_selection}),
\emph{summarizing reviews of places} (\cref{sec:system_case_study_hiro_summarize_reviews}),
and \emph{making action decisions} (\cref{sec:system_case_study_hiro_action_decision}).
These four components jointly enable Hiro to explore in our interactive embodied \geoul{environment} driven by his initial intention.

\subsection{Intention to Place Type}
\label{sec:system_case_study_hiro_intention_to_place_type}
Starting with a user-defined agent intention, Hiro first determines the type of place that could fulfill this intention using \llmul{GPT-4} and the following prompt:

\begin{quote}
{\small
\texttt{[Role]}\\
\texttt{You are PlaceSuggesterGPT, an expert in recommending types of places based on user-specified intentions.}\\

\texttt{[Task Description]}\\
\texttt{Given a user-specified intention, determine the type of "place" one should seek to fulfill the intention.
 Your response should be in the following JSON format:}\\
\texttt{\{"place": "Desired Place Type"\}}\\

\texttt{[Example]}\\
\texttt{Input:}
\texttt{"Intention: <buy a book>"}\\
\texttt{Output:}
\texttt{\{"place": "bookstore"\}}\\

\texttt{[Input]}\\
\texttt{Intention: <\{\textbf{agent\_intention\}}>}

\texttt{[Output]}\\
\texttt{Your recommended place type based on the user-specified intention, in the required JSON format:}\\
}
\end{quote}
\vspace{-1em}
Using this prompt with the intention
\begin{quote}
\textit{Hiro is hungry and looking for a place where he can try some good local food. He cannot handle spicy food.}
\end{quote}
returns the result
\begin{quote}
    \texttt{\{"place": "restaurant"\}}.
\end{quote}
The identified place type (here, \texttt{restaurant}) is extracted and set as the target category for Hiro's \cvul{open-world detector} during his exploration.

\subsection{Road Selection}
\label{sec:system_case_study_hiro_road_selection}
Whenever Hiro is at a crossroads, he determines the best road to follow using his \cvul{multi-modal LLM} and \llmul{GPT-4}. 
The primary goal of the road selection process is to identify the road most likely to lead to the desired place type that aligns with Hiro's intention. 
First, Hiro \geoul{fetches} the street view towards each potential road using the \virl environment. Then he utilizes his \cvul{multi-modal LLM} (such as InstructBLIP~\cite{dai2023instructblip} or LLaVA~\cite{liu2023visual}) to generate captions for each road using the following prompt:

\begin{quote}
{\small
\texttt{I am looking for a \textbf{\{place\_type\}}. Please detail information that might be helpful for me along this road:}
}
\end{quote}
Captions for each road are then formatted in the style of 
\begin{center}
\texttt{\textbf{\{road\_idx\}}: \textbf{\{road\_description\}}}
 \end{center}
and concatenated to form \texttt{all\_road\_descriptions}.
These road captions, along with Hiro's user-defined intention, are then fed into \llmul{GPT-4} to determine the most promising road to follow using the following prompt:

\begin{quote}
{\small
\texttt{[Role]}\\
\texttt{You are PathSelectorGPT, an expert in choosing the optimal road from multiple candidates based on a user-specified intention.}\\

\texttt{[Task Description]}\\
\texttt{Given an intention, the road previously traveled, and descriptions of available candidate roads, select the best road from the crossroad.
Your response must be in the following JSON format:}\\
\texttt{\{"idx": "Selected road index", "reason": "Justification for your selection"\}}

\texttt{[Example]}\\
\texttt{For the intention "find a grocery store", the road previously traveled as "1", and with candidates "2: Leads to residential area, 3: Leads to a shopping district", the output might be:
\{"idx": "3", "reason": "Road 3 leads to a shopping district which is more likely to have a grocery store."\}}

\texttt{[Input]}\\
\texttt{User Intention: <\textbf{\{agent\_intention\}}>}\\
\texttt{Road Descriptions: \textbf{\{all\_road\_descriptions\}}}\\
\texttt{Previously Traveled Road: Road \textbf{\{from\_road\_idx\}}}

\texttt{[Output]}\\
\texttt{Your chosen road index and the reasoning behind your selection, in the required JSON format:}
}
\end{quote}

We design such a two-stage captioning and decision-making pipeline for road selection because \cvul{Multi-modal LLMs} cannot process multiple images simultaneously. 
However, with the recent advancements of \cvul{GPT-4V}, it may be possible to perform road selection using several road images with a single prompt at once. 
Empirical findings suggest that \cvul{GPT-4V} yields more reasonable choices with the following prompt:
\begin{quote}
{\small
\texttt{[Role]}\\
\texttt{You are PathSelectorGPT, an expert in choosing the optimal road from multiple road images according to a user-specified intention.}\\

\texttt{[Task Description]}\\
\texttt{Given a set of road images, select the best road from the crossroad. Your answer must be in the following JSON format:}\\
\texttt{\{"idx": "Selected road index (start by 0)", "reason": "Justification for your selection"\}}\\

\texttt{[Input]}\\
\texttt{User Intention: <\textbf{\{agent\_intention\}}>}\\

\texttt{[Output]}\\
\texttt{Please answer with the road index and the reasoning behind your selection, in the required JSON format:}\\
}
\end{quote}
An example road selection response for the first crossroad selection in \cref{fig:agent_intention} is as follows:
\storytext{
    \includegraphics[height=0.9em,trim={0mm 1mm 0mm 1mm}]{figs/icons/star.png} {``idx'': ``0'', ``reason'': ``Choosing road 0 promises a genuine taste of local cuisine in a less commercialized setting. It's likely to have family-run eateries where I can request non-spicy dishes and savor authentic flavors. This road offers a tranquil dining atmosphere, allowing for a more engaged and leisurely culinary exploration among the locals.''}
}

\subsection{Summarize Place Reviews}
\label{sec:system_case_study_hiro_summarize_reviews}
When Hiro \cvul{discovers} a place in the street view imagery, he \geoul{retrieves} its corresponding name and Google reviews from the \virl environment. 
There is a complex algorithm behind projecting the 2D box on street view imagery to a concrete place in the real world, which is detailed in ``matching between object proposal and places'' of \cref{sec:benchmark_localization}.
After Hiro obtains these place reviews, he \llmul{summarizes} them into a place overview (to aid in decision-making) using the following prompt:

\begin{quote}
{\small
\texttt{[Role]}\\
\texttt{You are SummarizeGPT, skilled at condensing multiple reviews into a concise overview of a location.}\\

\texttt{[Task Description]}\\
\texttt{Given multiple reviews with ratings, craft a brief overview of the place. Your response should be in the following JSON format:}\\
\texttt{\{"summarization": "Concise description (limited to 80 words)"\}}\\

\texttt{[Example]}\\
\texttt{For reviews "Great ambiance but average food (Rating: 3)" and "Loved the decor, food could be better (Rating: 3.5)", the output might be:}\\
\texttt{\{"summarization": "The place boasts great ambiance and decor, but the food quality receives mixed reviews."\}}\\

\texttt{[Input]}\\
\texttt{Reviews: \textbf{\{all\_reviews\}}}\\

\texttt{[Output]}\\
\texttt{Your concise overview (max 80 words) based on the provided reviews, in the prescribed JSON format:}
}
\end{quote}

\subsection{Action Decision}
\label{sec:system_case_study_hiro_action_decision}
After obtaining the overview of the identified place, Hiro \llmul{decides} to visit the place or keep exploration using \llmul{GPT-4} and the following prompt:

\begin{quote}
{\small
\texttt{[Role]}\\
\texttt{You are ActionSelectorGPT, proficient in choosing the most appropriate action based on a user's background, intention, and an overview of a place.}\\

\texttt{[Task Description]}\\
\texttt{Evaluate the provided user background, intention, and place overview to select the most suitable action from the list. Your response should be in the following JSON format:}\\
\texttt{\{"action": "Selected Action", "reason": "Justification for your choice"\}}\\

\texttt{Possible actions:}\\
\texttt{- enter\_place(): Enter the designated place.}\\
\texttt{- continue(): Continue searching for another appropriate place.}\\

\texttt{[Example]}\\
\texttt{For the background "loves historical sites", intention "discover local history", and place overview "This is a 200-year-old preserved mansion", the output might be:}\\
\texttt{{"action": "enter\_place()", "reason": "The historical mansion aligns with the user's interest in historical sites."}}\\

\texttt{[Input]}\\
\texttt{User Background: <\textbf{\{background\}}>}\\
\texttt{User Intention: <\textbf{\{intention\}}>}\\
\texttt{Place Overview: <\textbf{\{place\_intro\}}>}\\

\texttt{[Output]}\\
\texttt{Your chosen action and the rationale behind your decision in the prescribed JSON format:}
}
\end{quote}
Hiro's exploration will continue if he decides to \texttt{continue()} and will terminate if he opts for \texttt{enter\_place()}.

\section{\virl Benchmarks: Details}
\label{sec:benchmark_details}

\subsection{\virl Places: Detection (Details)}
\label{sec:benchmark_details_localization}

\begin{figure}[h]
    \centering
    \includegraphics[width=1.0\linewidth]{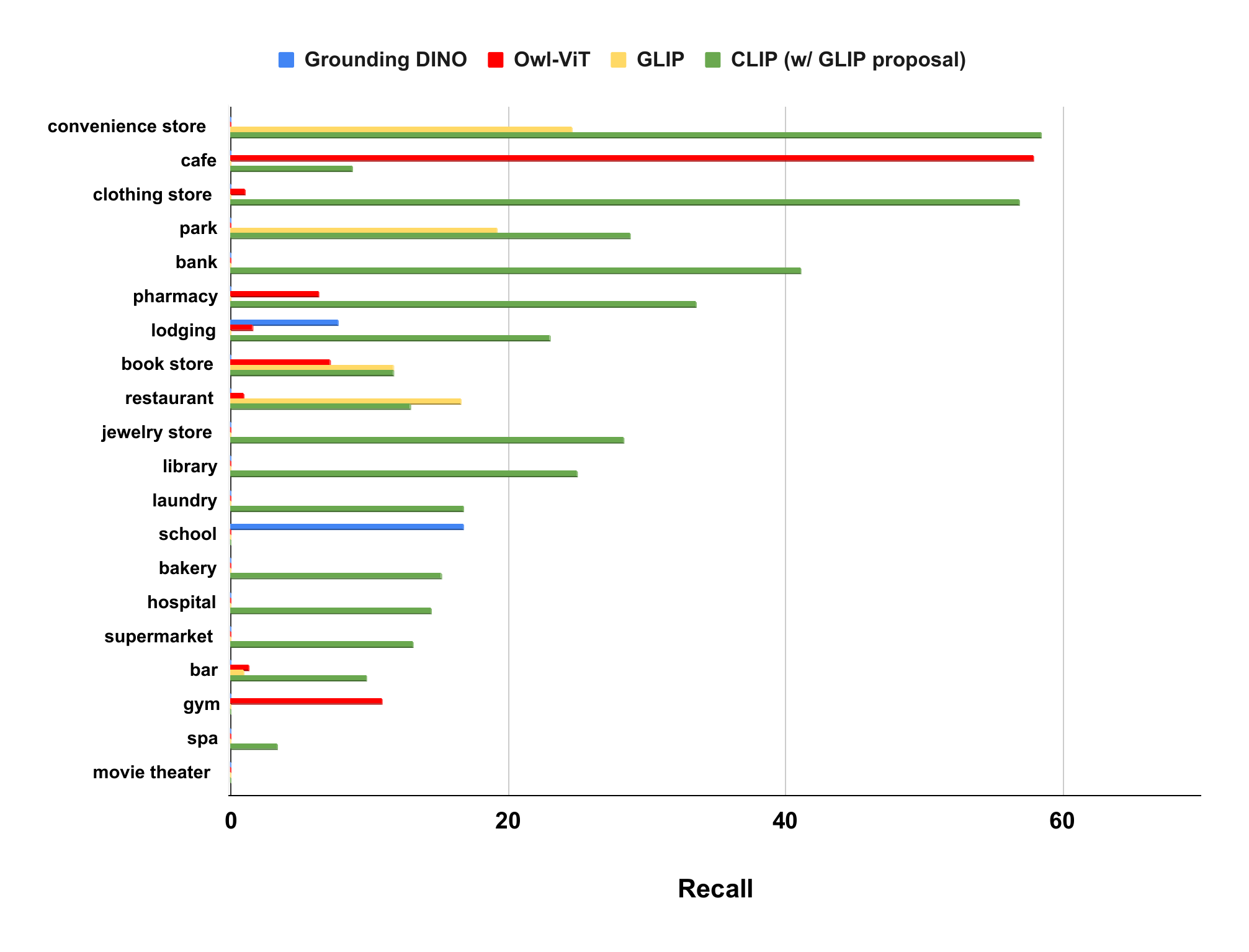}
    \vspace{-1.0cm}
    \caption{Recalls in \virlplace detection}
    \vspace{-0.1cm}
    \label{fig:bm_det_accuracy}
\end{figure}

\paragraph{All category results.} Due to the page limit of the main paper, we only present the results of 10 categories in \cref{tab:benchmark_det_res}. Here, we present the place recall for all 20 categories in \cref{fig:bm_det_accuracy}.

\begin{figure*}[h]
    \centering
    \includegraphics[width=1.0\linewidth]{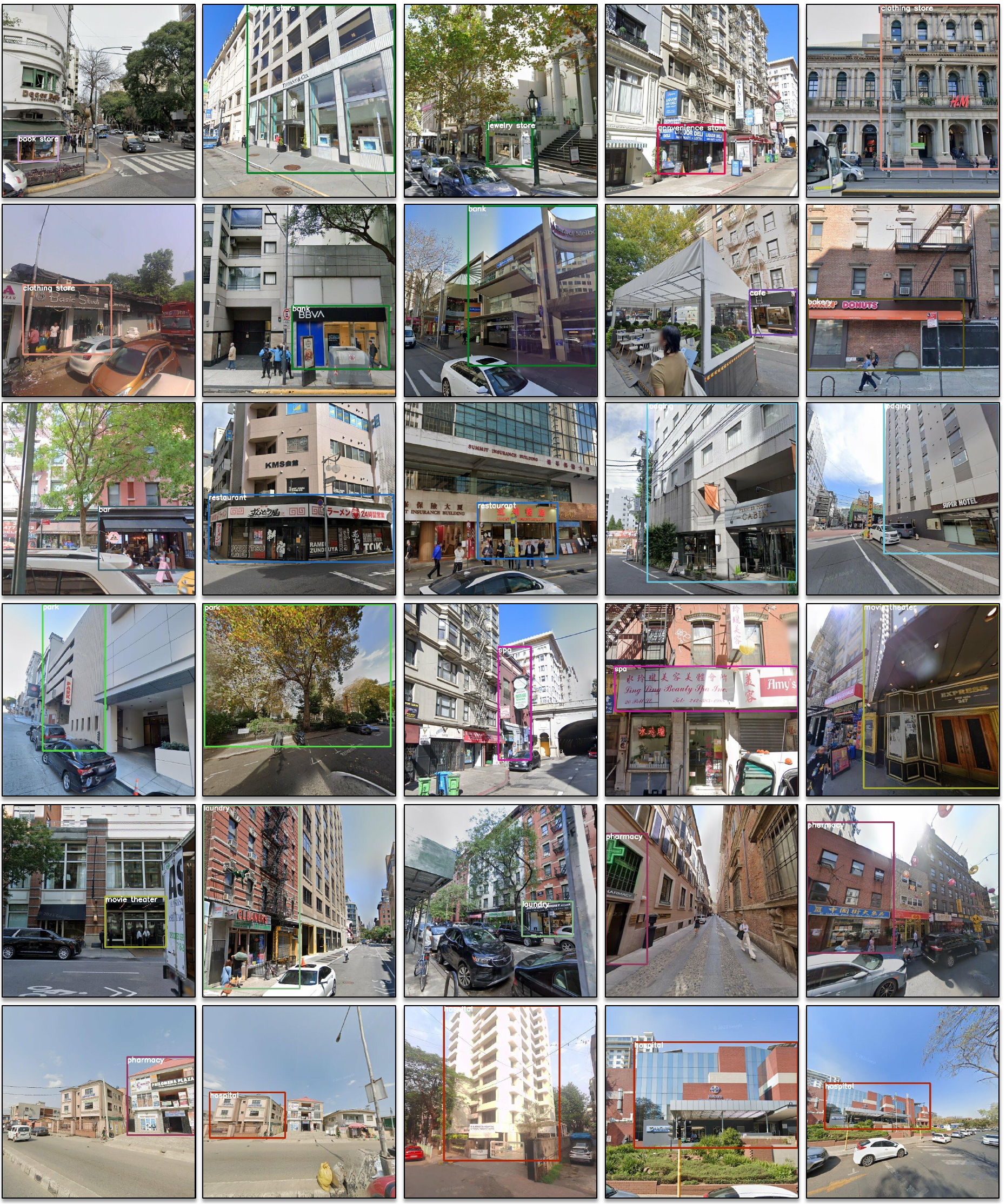}
    \vspace{-0.3cm}
    \caption{Samples of \virlplace detection using CLIP (w/ GLIP proposals).}
    \vspace{-0.1cm}
    \label{fig:bm_det_examples}
\end{figure*}

\paragraph{Example illustrations.} To facilitate the understanding of \virlplace detection benchmark, we present some examples of CLIP (w/ GLIP proposals) in \cref{fig:bm_det_examples}. 

\begin{figure}[h]
    \centering
    \includegraphics[width=0.8\linewidth]{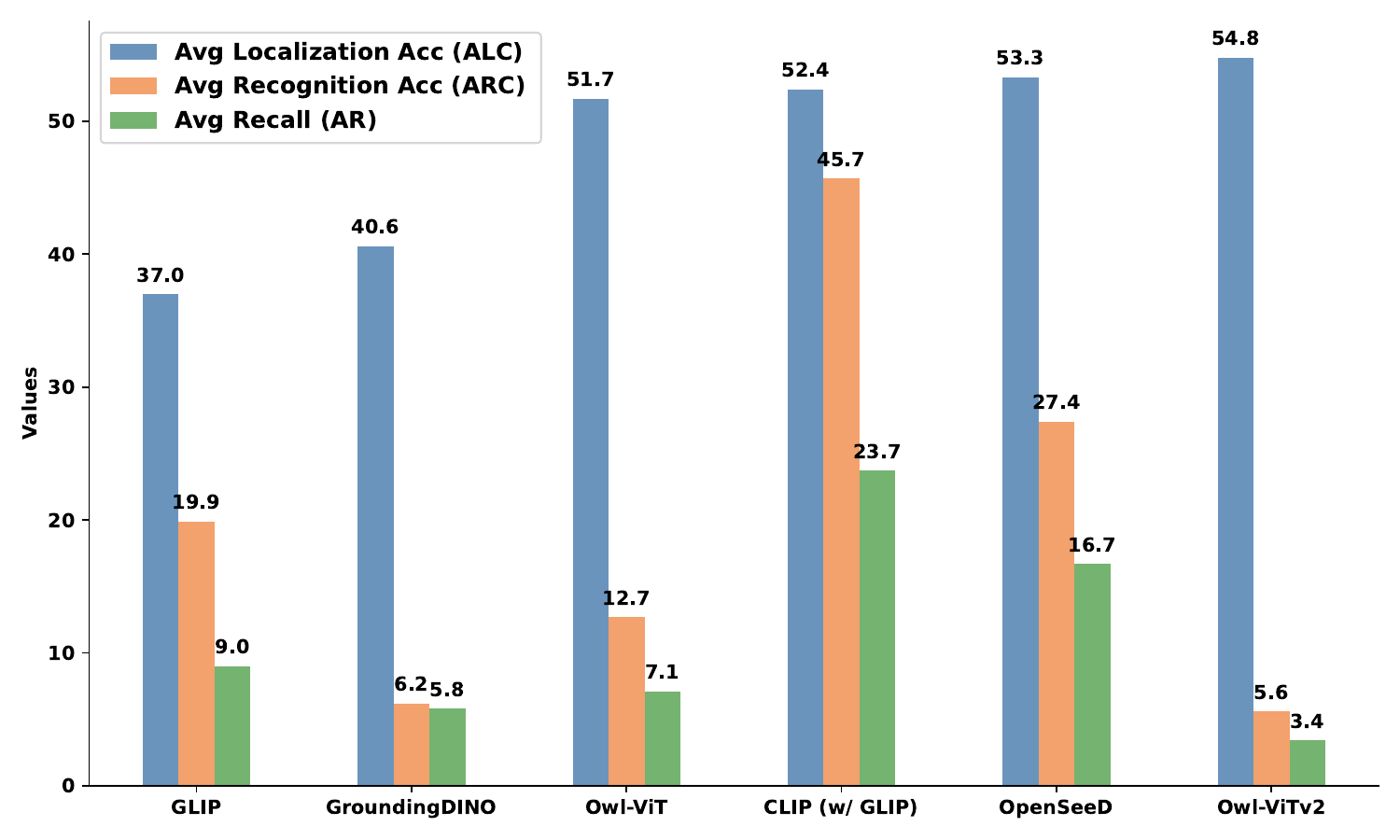}
    \vspace{-0.3cm}
    \caption{Error diagnosis in \virlplace Detection}
    \vspace{-0.1cm}
    \label{fig:bm_det_error_analysis}
\end{figure}

\paragraph{Error Diagnosis of Detectors.} We conduct error diagnosis for detectors in the \virlplace detection benchmark. We examine two error types: localization error and classification error. As depicted in \cref{fig:bm_det_error_analysis}, the primary challenge arises in classification, where detectors struggle to assign correct labels, despite having accurate object proposals.

\subsection{\virl Places: Recognition and VQA (Details)}
\label{sec:benchmark_details_rec_vqa}

\begin{figure}
    \centering
    \includegraphics[width=1.0\linewidth]{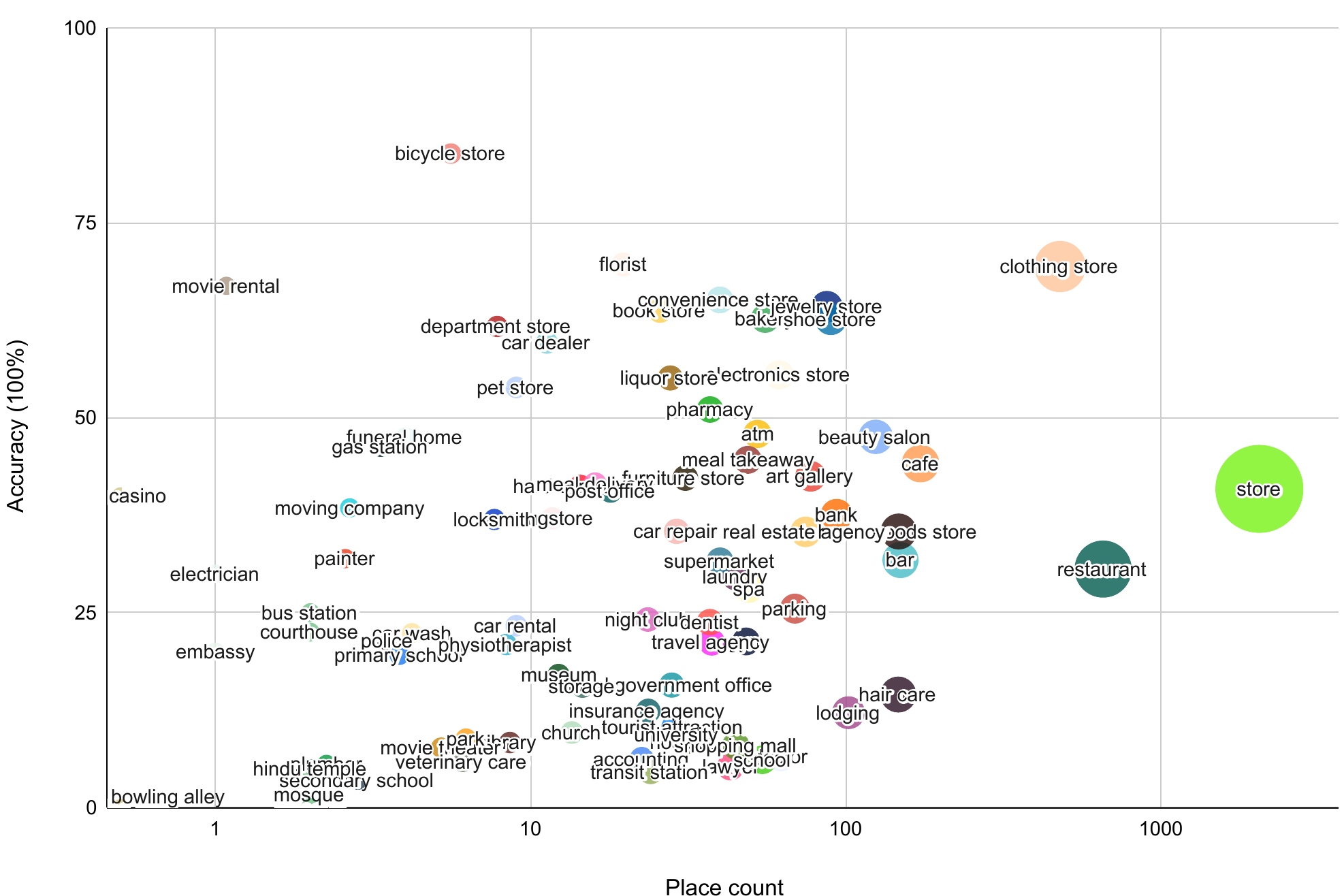}
    \caption{Category-wise accuracy and numbers for \virl Place Recognition benchmark.
    }
    \label{fig:benchmark_rec_place_type_results}
\end{figure}

\paragraph{Place types performance for recognition.} In Figure \ref{fig:benchmark_rec_place_type_results}, we present the averaged accuracy for each place type across 10 benchmarked vision models. The size and the x-axis position of each bubble correspond to the number of places within each type. A clear trend emerges: accuracy tends to correlate with the frequency. Common categories such as \texttt{clothing store}, \texttt{cafe} exhibit higher accuracy, whereas vision models often struggle with infrequent place types like \texttt{bowling alley} or \texttt{mosque}. 

\begin{figure}[h!]
    \centering
    \includegraphics[width=1.0\linewidth]{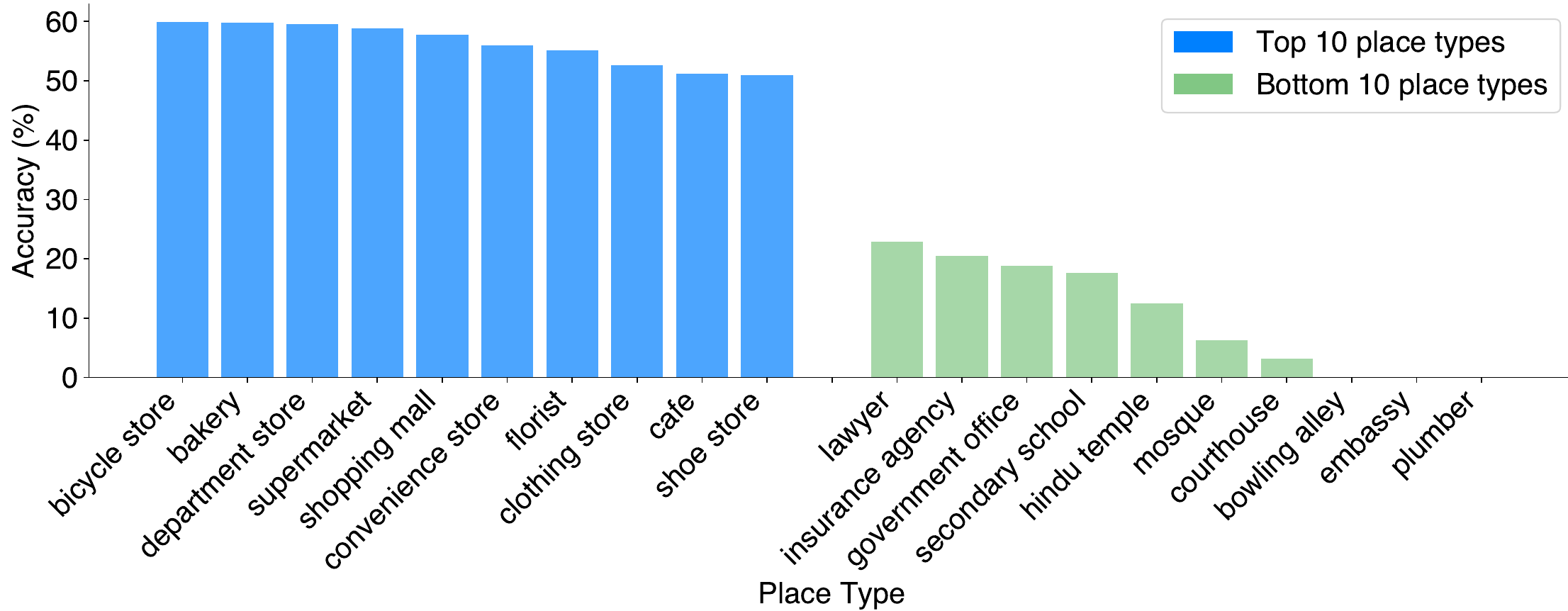}
    \vspace{-0.2cm}
    \caption{Top-10 and bottom-10 place types averaged on four vision models of \virl Place VQA.}
    \vspace{-0.1cm}
    \label{fig:benchmark_vqa_place_type_results}
\end{figure}

\paragraph{Place types performance for VQA.} The place types performance of the \virl place VQA in \cref{fig:benchmark_vqa_place_type_results} further verifies the correlation between accuracy and frequency from a human intention perspective. The top-10 categories are closely aligned with the most common human activities, purchasing and dining. In contrast, the bottom-10 place types relate to places that are less frequently encountered and serve a more diverse purpose, such as \texttt{mosque}, \texttt{plumber} and \texttt{embassy}.

\paragraph{Consistency Analysis of \virlplace VQA Results.} Here, we study the \textit{sensitivity of MLLM to the order of QA options} in VQA. 
As shown in \cref{tab:vqa_consistency_analysis}, advanced MLLMs still exhibit 4.5\%-10.6\% mAcc drop with circular evaluation. This highlights that MLLMs are still sensitive to the order of QA options presented. 

\begin{table}[h!]
\centering
\begin{small}
\setlength\tabcolsep{3pt}
\scalebox{0.86}{
    \begin{tabular}{l|ccc}
    \toprule
    \textbf{Method} & InternVL-1.5~\cite{chen2024internvl1.5} & GPT-4V~\cite{openai2023gpt} & Qwen-VL-Max~\cite{bai2023qwen} \\
    \hline
    mAcc (w/ circular) & 77.6 & 77.1 & 74.3 \\
    mAcc (w/o circular) & 82.1 & 83.1 & 84.9 \\
    \textbf{mAcc drop} & \textbf{\textcolor{red}{-4.5}} & \textbf{\textcolor{red}{-6.0}} & \textbf{\textcolor{red}{-10.6}} \\
    \bottomrule
    \end{tabular}
}
\end{small}
\caption{MLLM consistency analysis on \virlplace VQA benchmark.}
\vspace{-0.3cm}
\label{tab:vqa_consistency_analysis}
\end{table}

\subsection{\virl Vision-Language Navigation (Details)}
\label{sec:benchmark_details_vln}

\paragraph{Navigation pipeline.} 
As mentioned in \cref{sec:ling_details}, our VLN pipeline is similar to \cite{schumann2023velma}, however, our benchmark offers greater scalability through the worldwide \virl platform and an automated data collection pipeline, as opposed to the manual annotation of a specific region. 
Furthermore, our benchmark emphasizes the analysis of the \emph{vision} component in the VLN pipeline, as opposed to \cite{schumann2023velma}, which aims to enhance performance on existing VLN datasets using LLMs.

\paragraph{Implementation Details.} Here, we introduce the implementation details for LLaVA-1.5~\cite{liu2023improvedllava} and PP-OCR~\cite{du2009pp} (+ GPT-3.5). 
For LLaVA-1.5~\cite{liu2023improvedllava}, we transform the landmark recognition task to a \emph{multiple choice VQA} problem, asking
\begin{quote}
    \texttt{Which of the following landmarks can be identified with a high degree of confidence?}
\end{quote}
The VQA options include all potential landmarks mentioned in the route description, along with a ``\texttt{None of above}'' choice. 
The model's response to this question is then parsed as the landmark observation.

For PP-OCR~\cite{du2009pp} (+ GPT-3.5), we first extract all recognized text using PP-OCR~\cite{du2009pp} for each street view image. Then, GPT-3.5~\cite{schulman2022chatgpt} determines the presence of each landmark in this street view image, jointly considering the OCR text and landmark name.

\vspace{-0.2cm}
\paragraph{Full set results.} Apart from the mini-set results presented in \cref{sec:benchmark_vln}, we also provide the full set results of Oracle and CLIP (L/14@336px) in \cref{tab:vln_results_fullset}. 
The Oracle results, interestingly, do not achieve a 100\% success rate, due to incorrect decisions made by the LLM at stop positions. This is evidenced by the high arrival ratio and low reaction accuracy at stop positions. Empirically, we observe that the LLM occasionally decides to keep moving, despite clear destination indications in the observations. 

When we substitute the map in oracle with the CLIP model to gather landmark observations from street view imagery, we observe a significant drop in the success rate, due to the inevitable model prediction errors. 
To improve the success rate in VLN, we can focus on two important factors: ($i$) designing better vision models; ($ii$) developing LLMs and prompt techniques that are robust to vision-related noise. Especially, our empirical findings suggest that sophisticated prompt designs significantly improve the robustness of LLMs to visual observation noise.

\begin{table}[h!]
\centering
\begin{small}
\setlength\tabcolsep{3pt}
\scalebox{0.92}{
    \begin{tabular}{lcccccc}
        \toprule
            \multirow{2}{*}{\textbf{Method}} & \multicolumn{1}{c}{\textbf{}} & \multicolumn{1}{c}{\textbf{Start}} & \multicolumn{2}{c}{\textbf{Intersection}} & \multicolumn{2}{c}{\textbf{Stop}}\\
            \cmidrule{3-7}
            & \textbf{Success} & \textbf{Reac} & \textbf{Arr} & \textbf{Reac} & \textbf{Arr} & \textbf{Reac}\\
        \midrule
        \cellcolor{mygray}Oracle (No Vision) & \cellcolor{mygray}0.88  & \cellcolor{mygray}1.0 & \cellcolor{mygray}0.95 & \cellcolor{mygray}0.99 & \cellcolor{mygray}0.96 & \cellcolor{mygray}0.88\\
        \midrule
        CLIP (L/14@336px) & 0.22  & 0.84 & 0.66 & 0.90 & 0.61 & 0.22 \\ 
        \bottomrule
    \end{tabular}
}
\end{small}
\caption{Results of \virl VLN-full.}
\label{tab:vln_results_fullset}
\end{table}

\end{document}